\documentclass[10pt,a4paper,twocolumn]{article}
\usepackage[T1]{fontenc}
\usepackage[utf8]{inputenc}
\usepackage[english]{babel}
\usepackage{stix2}
\usepackage[left=16mm,right=16mm,top=18mm,bottom=20mm,columnsep=6mm]{geometry}
\usepackage{microtype}
\usepackage{amsmath}
\usepackage{graphicx}
\usepackage{array,booktabs,colortbl}
\usepackage{xcolor,textcomp,steinmetz}
\usepackage{etoolbox}
\usepackage{stfloats}
\usepackage[font=footnotesize,labelfont=bf,labelsep=period,justification=justified,singlelinecheck=false]{caption}
\usepackage[numbers,sort&compress]{natbib}
\usepackage{xurl}
\usepackage[unicode,colorlinks=true,linkcolor=black,citecolor=black,urlcolor=blue,bookmarksopen=true]{hyperref}
\usepackage{bookmark}
\graphicspath{{figures/}}
\AtBeginEnvironment{tabular*}{\footnotesize\setlength{\tabcolsep}{4pt}}
\newenvironment{tablenotes}{\par\vspace{4pt}\footnotesize\noindent\ignorespaces}{\par}

\DeclareUnicodeCharacter{03BC}{\textmu}                   
\DeclareUnicodeCharacter{03B5}{\ensuremath{\varepsilon}}  
\DeclareUnicodeCharacter{0394}{\ensuremath{\Delta}}
\DeclareUnicodeCharacter{03C4}{\ensuremath{\tau}}
\DeclareUnicodeCharacter{03BB}{\ensuremath{\lambda}}
\DeclareUnicodeCharacter{207B}{\textsuperscript{\textminus}}
\DeclareUnicodeCharacter{00B9}{\textsuperscript{1}}
\DeclareUnicodeCharacter{00B2}{\textsuperscript{2}}
\DeclareUnicodeCharacter{00B3}{\textsuperscript{3}}
\DeclareUnicodeCharacter{2080}{\textsubscript{0}}
\DeclareUnicodeCharacter{2083}{\textsubscript{3}}
\DeclareUnicodeCharacter{2010}{-}
\DeclareUnicodeCharacter{2212}{\textminus}
\DeclareUnicodeCharacter{2192}{\ensuremath{\rightarrow}}  

\newcommand{\capline}[1]{ #1}

\newcommand{\adrrfig}[4]{%
  \begin{figure*}[!tbp]
  \centering
  \includegraphics[width=\textwidth]{#2}
  \caption[#3]{#3#4}\label{#1}
  \end{figure*}}

\hypersetup{
  pdftitle={Non-Imaging Optical Tactile Sensors for Robotics: Transduction Principles, System Integration, and Robotic Applications},
  pdfauthor={Hao Wu; Yu Feng; Haotian Guo; Huichan Zhao; Changsheng Wu; Jianshu Zhou},
  pdfsubject={Review of non-imaging optical tactile sensing for robotics},
  pdfkeywords={optical tactile sensing, non-imaging sensors, robotic perception, fiber-optic sensors},
  pdflang={en-US}
}
\begin{document}
\twocolumn[{
\begin{center}
{\LARGE\bfseries Non-Imaging Optical Tactile Sensors for Robotics:\\[3pt]
Transduction Principles, System Integration, and Robotic Applications\par}
\vspace{1em}
{\large Hao Wu\textsuperscript{1,\dag}, Yu Feng\textsuperscript{1,\dag}, Haotian Guo\textsuperscript{1},
Huichan Zhao\textsuperscript{2}, Changsheng Wu\textsuperscript{3}, Jianshu Zhou\textsuperscript{1,*}\par}
\vspace{.65em}
{\small
\textsuperscript{1}Department of Mechanical Engineering, National University of Singapore, Singapore\\
\textsuperscript{2}Department of Mechanical Engineering, Tsinghua University, Beijing, China\\
\textsuperscript{3}Department of Materials Science and Engineering, National University of Singapore, Singapore\par}
\vspace{.5em}
{\small\textsuperscript{\dag}Hao Wu and Yu Feng contributed equally to this work.\\
\textsuperscript{*}Correspondence: Jianshu Zhou (\href{mailto:jianshuzhou@nus.edu.sg}{jianshuzhou@nus.edu.sg})\par}
\end{center}
\vspace{.5em}
{\small\noindent\textbf{Abstract.} Non-imaging optical tactile sensors encode contact through changes in optical intensity, wavelength, spectra, or photon counts, offering diverse architectures for robotic touch. Their usefulness depends on whether contact variables can be reliably distinguished from coupled mechanical and environmental effects, and whether the resulting information can be delivered in time to guide action. This review develops a unified framework linking mechanical load transmission, optical encoding, readout, and tactile inference to robotic task requirements. It examines reflective and waveguide-based devices, micro/nanofibers, fiber gratings, interferometric and resonant structures, and mechanoluminescent composites. Three considerations emerge across these approaches. First, high optical sensitivity supports accurate tactile estimation only when target variables remain identifiable under the relevant contact and calibration conditions. Second, spatial coverage and feedback timing must be assessed at the system level, accounting for multiplexing, acquisition, demodulation, and inference. Third, device characterization, offline recognition, and closed-loop robotic performance provide distinct levels of evidence and require different evaluation criteria. Representative studies illustrate how these considerations shape force estimation, contact localization, slip correction and distributed interaction. This review concludes with priorities for calibration transfer, reliable interfaces, scalable readout, and task-based validation, providing a basis for selecting and developing non-imaging optical tactile architectures for robotics.\par}
\vspace{.7em}
{\small\noindent\textbf{Keywords:} tactile sensors; optoelectronic tactile sensors; soft sensors; robotic perception; robotic touch; grasping and manipulation\par}
\vspace{1.5em}
}]

\section{Introduction}\label{sec:1}

During contact, robots sense contact loads and object motion to guide their actions. When grasping fragile objects, grip force must be regulated to maintain a stable grasp without excessive compression. When an object begins to slip, the robot must correct its action promptly~\cite{wang2026tangentially,lee2026bioinspired}. When exploring soft tissues or compliant materials, their mechanical responses must be assessed by considering both the applied load and the indentation process. Bending of soft fingers can also affect sensor outputs, making it necessary to distinguish external contact from self-deformation for accurate interpretation of tactile signals. These requirements have driven the development of tactile devices for robotic end effectors~\cite{zhao2016optoelectronically,massari2022functional,ma2026low,jiang2024multifunctional,zhou2026artificial}, and they broaden as dexterous hands and soft grippers take on more varied contact tasks~\cite{guo2025enabling,huang2026bioinspired,boron2026stochastic,feng2026gecko,zhou2025dexterous,zhou202050,cao2024design,zhou2026everything,zhou2020adaptive}. 

Tactile sensors acquire contact signals through different transduction mechanisms and readout methods. Common sensing principles for soft robotic tactile skins include resistive, capacitive, magnetic, pneumatic and optical transduction~\cite{roberts2021soft}. Electronic skins for prosthetics have pursued mechanical durability, stretchability, biodegradability and the sensing of diverse stimuli over large areas~\cite{chortos2016pursuing,wang2015recent}. Capacitive sensors generate capacitance signals through changes in electrode spacing, effective area or dielectric properties, whereas resistive sensors convert deformation into resistance signals through changes in conductive pathways or contact resistance~\cite{song2026flexible,cheng2026understanding,choi2024stretchable,yang2025dual,feng2025enhancing}. Multimodal electrical sensors read different stimuli separately through decoupled sensing mechanisms or structural arrangements~\cite{feng2026intrinsic,yang2022multimodal,zhu2024multifunctional}. Magnetic tactile sensing uses deformation of a magnetic elastomer or displacement of an embedded magnet to alter the local magnetic field, which is then measured by a magnetic field detector~\cite{yan2021soft,man2024magnetic,huang2026miniaturized,feng2026bioinspired}. 

Optical tactile sensing converts contact into changes in light propagation or emission~\cite{zhao2016optoelectronically,yang2026recent}. In vision-based tactile sensing, a camera records contact-layer topography or marker displacement, from which tactile information is reconstructed using images~\cite{yuan2017gelsight,wang2026flexible,wang2026endotac}. Such sensors can be grouped into marker-based and intensity-based transduction~\cite{li2025classification}, and an open-source family of dual-material 3D-printed skins tracks internal pins with a camera~\cite{wardcherrier2018tactip}. Camera readout gives these sensors high spatial resolution, but the hardware can be bulky compared with thin-film sensors, and the response speed is generally limited by the camera frame rate, mostly 30–90 frames/s, and by the computation of the tactile model~\cite{li2025classification}. Photodetector arrays used in place of cameras resolve only several millimeters but sample at 1 kHz or more~\cite{li2025classification}. This review focuses on non-imaging optical tactile sensing, which acquires contact information by measuring optical intensity, wavelength, spectra or the temporal characteristics of light emission (Figure~\ref{fig:1})~\cite{zhao2016optoelectronically,heo2006tactile,luo2026self}. Vision-based tactile sensing that uses images of the contact surface as tactile input falls outside the scope of this review. Line-scan cameras that sample fixed waveguide output ports and cameras used solely for calibration do not acquire images of the contact surface and are therefore classified as non-imaging readout, whereas charge-coupled device (CCD) prototypes that image the fingertip contact surface are excluded~\cite{khamis2019novel,hu2023polymer,maekawa1993finger}. 

Optical tactile sensing allows contact structures and signal readout locations to be designed separately according to installation requirements. Optical fibers connect sensitive regions to light sources and detectors, allowing some electronic components to be located away from the contact area~\cite{guo2026intelligent}. Soft waveguides and flexibly encapsulated optical fibers can be arranged along fingers or curved skins to transmit optical signals while accommodating surface deformation, making them suitable for compliant fingers, underwater contact and remote detection~\cite{zhao2016optoelectronically,massari2022functional,teeple2018soft}. Optical-fiber transmission sections convey contact information through optical signals rather than conductive currents, enabling electrical isolation and reducing the influence of electromagnetic interference on signal transmission. For distributed suction cups or local interaction, discrete emitters, photodetectors and integrated optoelectronic chips can also form local readout units, allowing the sensing layout to accommodate the robot's installation space and contact requirements~\cite{deldottore2026peripheral,shi2026multi}.

Optical encoding also provides different means of observation for multipoint and multiparameter sensing. Multiple fiber Bragg gratings can be wavelength-division multiplexed along a single fiber, connecting several sensitive regions through one transmission path. Changes in the color and optical intensity of soft lightguides can encode deformation locations, and distinct resonance responses can distinguish contact force from temperature~\cite{heo2006tactile,bai2020stretchable,shang2023soft}. These observables can support tactile coverage, differentiation between self-deformation and external contact, and multivariable feedback.

Optical tactile sensing has developed along imaging and non-imaging routes for about four decades (Figure~\ref{fig:timeline}). In the 1980s, two lines developed in parallel. One imaged the contact: frustrated total internal reflection at a light-conducting plate or waveguide formed a contact imprint that a solid-state camera digitized~\cite{king1985tactile,begej1988planar}, and coherent fiber bundles carried light reflected from a deformable skin to a television camera~\cite{schneiter1984optical}. This line later led to retrographic sensing with a reflective elastomer skin~\cite{johnson2009retrographic} and to the vision-based sensors outside the scope of this review. The other line read light intensity element by element. A stacked light-emitting diode (LED) matrix, photodetector matrix and pressure-receiving layer formed a 15 × 15 array with 2 mm pitch~\cite{nomura1985two}. Fibers arranged in four overlapping layers served both as microbend force sensors and as signal paths~\cite{jenstrom1989fiber}. In a modular sensor for a multifingered hand, the unit was a cantilever that reflected light from an emitting fiber into a receiving fiber~\cite{ristic1989analysis}. Miniaturization followed. A finger-shaped waveguide first read with a CCD camera~\cite{nakao1990finger} was converted to a position-sensitive detector that outputs the centroid of the scattered light~\cite{maekawa1993finger}, and a surface-emitting laser fed back by an external mirror served as both light source and interferometer~\cite{yamamoto1995optical}. In the 2000s, a single LED matrix acted as both emitter and detector~\cite{rossiter2005novel}, modules of photo-reflectors under urethane foam covered the curved surfaces of a humanoid robot~\cite{ohmura2006conformable}, and fiber Bragg grating arrays and gratings embedded in robot end effectors introduced wavelength encoding~\cite{heo2006tactile,park2009exoskeletal}. Prismatic-tip fiber arrays for dexterous hands, transparent and polymer-waveguide skins, and soft waveguides in prosthetic and deep-sea fingers then brought the approach to compliant robots~\cite{ataollahi2010tactile,ramuz2012transparent,yun2014polymer,zhao2016optoelectronically,teeple2018soft}. Since 2019, three-dimensional lightguide lattices, pillar arrays read by quadrant photodiodes, color-encoded stretchable lightguides, Ruffini-type grating skins, micro/nanofiber multi-axis sensors, mechanoluminescent fibers and an octopus-inspired soft arm have extended the readout to networks, spectra and emitted light~\cite{xu2019optical,khamis2019novel,bai2020stretchable,massari2022functional,xie2024optical,luo2026self,deldottore2026peripheral}.

Existing reviews have discussed the role of touch in manipulation from the perspectives of tactile integration in dexterous hands, the hierarchy of tactile information from raw signals to contact, object and action levels, and trends in robotic manipulation~\cite{kappassov2015tactile,li2020review,billard2019trends}. Early surveys already listed optical transduction among tactile sensing principles~\cite{nicholls1989survey,dahiya2010tactile}, with freedom from electrical interference and easy separation of the sensor from its processing electronics as its attractions~\cite{nicholls1989survey}. Reviews specific to fiber-optic tactile sensing have since covered modulation mechanisms, flexible packaging, array integration and intelligent algorithms~\cite{sun2026enabling}, as well as principles, fabrication and data processing~\cite{zhu2025advancements}. Others address sensors based on fiber Bragg gratings, micro/nanofibers, specklegrams and Fabry–Pérot interferometers~\cite{lyu2025fiber}, and micro/nanofiber tactile sensors together with soft actuators~\cite{zhang2024optical}.

This review organizes non-imaging optical tactile sensing around a central question: how can contact states be encoded into optical observations, reliably inferred under changing conditions, and used to guide robotic action? We connect mechanical load transmission, optical encoding, readout architecture and tactile inference to examine this question across device families ranging from reflective and waveguide-based sensors to fiber gratings, interferometric structures and mechanoluminescent devices~\cite{heo2006tactile,luo2026self,jenstrom1989fiber,xu2019optical}. The framework emphasizes three considerations: the identifiability of contact variables, the system-level constraints on spatial coverage and feedback timing, and the evidence connecting tactile measurements to robotic task outcomes. These considerations guide the analysis of transduction principles and device design, system integration, information decoupling and reconstruction, performance characteristics and robotic applications, and provide a basis for identifying priorities for practical deployment. Figure~\ref{fig:1} provides an overview of the transduction mechanisms and robotic applications examined within this framework.

\adrrfig{fig:1}{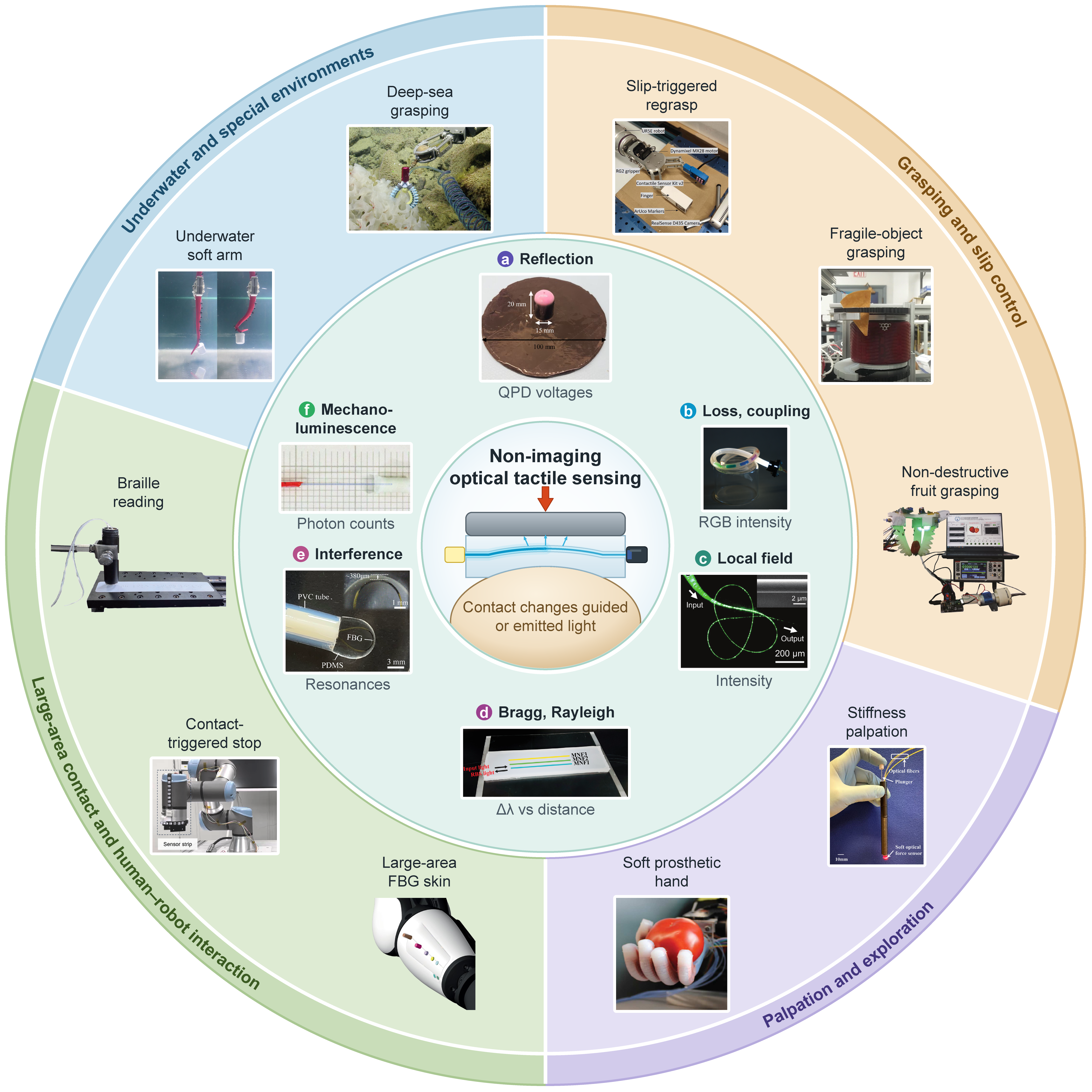}%
  {Overview of non-imaging optical tactile sensing for robots.}%
  {%
\capline{Center: contact modulates guided or emitted light (schematic). Inner ring: six transduction mechanisms with representative devices (a–f, see Figure~\ref{fig:2}); the gray line under each photograph names the quantity read. Outer ring: robotic applications in four groups (Section~\ref{sec:6}). The deep-sea photograph shows the gripper platform on earlier expeditions; the study's waveguide sensors were tested in simulated deep-sea conditions~\cite{teeple2018soft}.}
\capline{Inner ring: a) Reproduced with permission~\cite{khamis2019novel}. Copyright 2019, Elsevier. b) Reproduced with permission~\cite{bai2020stretchable}. Copyright 2020, The American Association for the Advancement of Science. c) Reproduced under the terms of the CC BY license~\cite{xie2024optical}. Copyright 2024, The Authors, published by Wiley-VCH. d) Reproduced with permission~\cite{li2026continuous}. Copyright 2026, Wiley-VCH. e) Reproduced with permission~\cite{shang2023soft}. Copyright 2023, American Chemical Society. f) Reproduced with permission~\cite{luo2026self}. Copyright 2025, Wiley-VCH. Outer ring, clockwise from the top: Slip-triggered regrasp: Reproduced with permission~\cite{wang2024robust}. Copyright 2023, IEEE. Fragile-object grasping: Reproduced under the terms of the CC BY license~\cite{massari2019tactile}. Copyright 2019, The Authors, published by Frontiers. Non-destructive fruit grasping: Adapted under the terms of the CC BY 4.0 license~\cite{fan2026intelligent}. Copyright 2026, The Authors, published by Springer Nature. Stiffness palpation: Reproduced under the terms of the CC BY license~\cite{ma2026low}. Copyright 2026, The Authors, published by Wiley-VCH. Soft prosthetic hand: Reproduced with permission~\cite{zhao2016optoelectronically}. Copyright 2016, The Authors, some rights reserved; exclusive licensee AAAS. Large-area FBG skin: Adapted under the terms of the CC BY 4.0 license~\cite{massari2022functional}. Copyright 2022, The Authors, published by Springer Nature. Contact-triggered stop: Reproduced with permission~\cite{kim2026modular}. Copyright 2026, Elsevier. Braille reading: Adapted under the terms of the Optica Open Access Publishing Agreement~\cite{wang2025optical}. Copyright 2025, Optica Publishing Group. Underwater soft arm: Reproduced under the terms of the CC BY 4.0 license~\cite{deldottore2026peripheral}. Copyright 2026, The Authors, published by Springer Nature. Deep-sea grasping: Reproduced with permission~\cite{teeple2018soft}. Copyright 2018, IEEE.}%
  }

\adrrfig{fig:timeline}{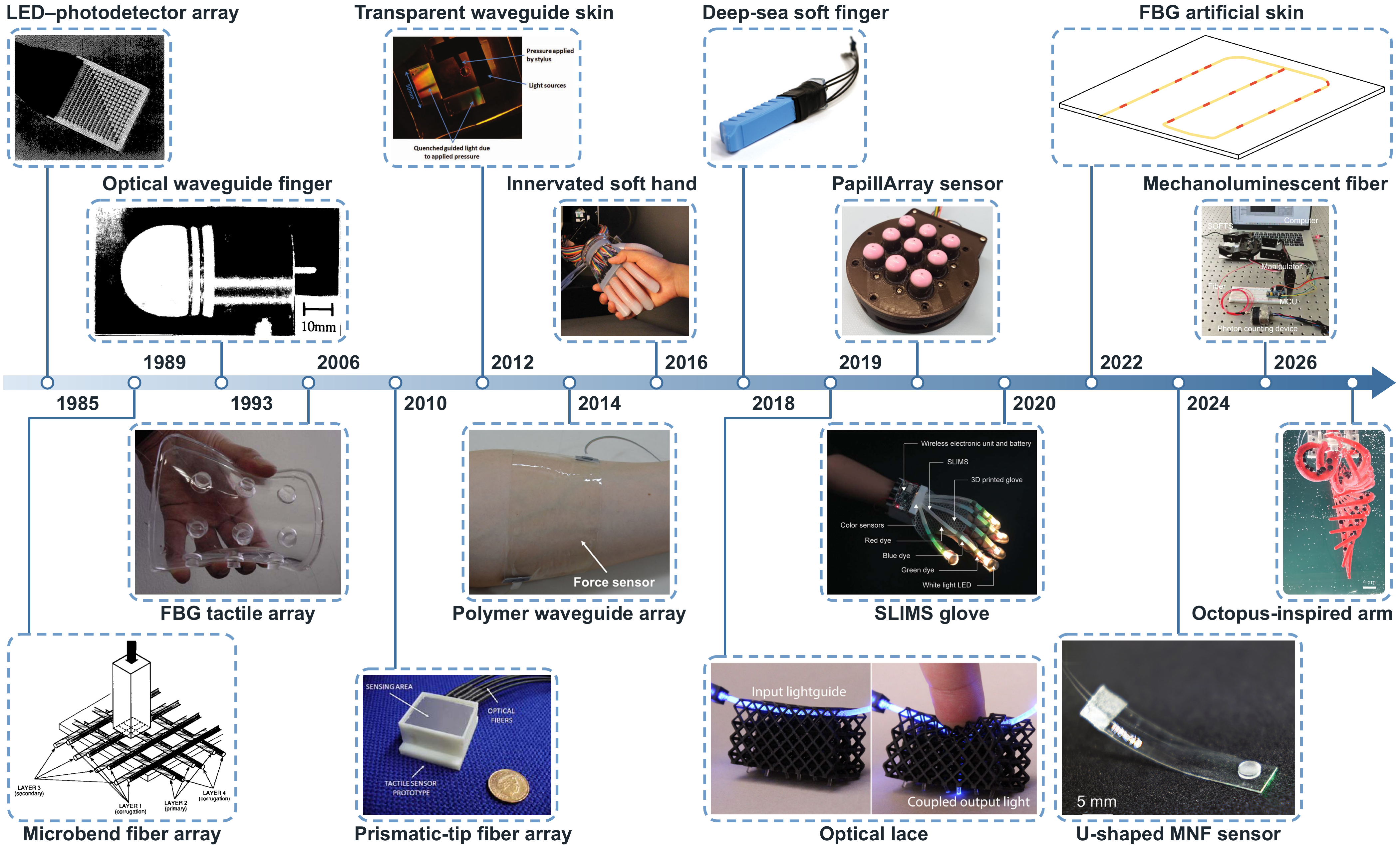}%
  {Timeline of non-imaging optical tactile sensors, 1985--2026.}%
  {%
\capline{Each milestone is shown with an image from the cited work and the name of the device. Milestones are evenly spaced, so the axis is not to scale. The 1993 waveguide finger is the miniaturized version read by a position-sensitive detector. FBG, fiber Bragg grating; MNF, micro/nanofiber; SLIMS, stretchable lightguide for multimodal sensing.}
\capline{In chronological order: LED--photodetector array: Reproduced with permission~\cite{nomura1985two}. Copyright 1985, IEEE. Microbend fiber array: Reproduced with permission~\cite{jenstrom1989fiber}. Copyright 1989, Elsevier. Optical waveguide finger: Reproduced with permission~\cite{maekawa1993finger}. Copyright 1993, IEEE. FBG tactile array: Reproduced with permission~\cite{heo2006tactile}. Copyright 2005, Elsevier. Prismatic-tip fiber array: Reproduced with permission~\cite{ataollahi2010tactile}. Copyright 2010, IEEE. Transparent waveguide skin: Reproduced with permission~\cite{ramuz2012transparent}. Copyright 2012, Wiley-VCH. Polymer waveguide array: Reproduced with permission~\cite{yun2014polymer}. Copyright 2014, Wiley-VCH. Innervated soft hand: Reproduced with permission~\cite{zhao2016optoelectronically}. Copyright 2016, The Authors, some rights reserved; exclusive licensee AAAS. Deep-sea soft finger: Reproduced with permission~\cite{teeple2018soft}. Copyright 2018, IEEE. Optical lace: Reproduced with permission~\cite{xu2019optical}. Copyright 2019, The American Association for the Advancement of Science. PapillArray sensor: Reproduced with permission~\cite{khamis2019novel}. Copyright 2019, Elsevier. SLIMS glove: Reproduced with permission~\cite{bai2020stretchable}. Copyright 2020, The American Association for the Advancement of Science. FBG artificial skin: Reproduced under the terms of the CC BY 4.0 license~\cite{massari2022functional}. Copyright 2022, The Authors, published by Springer Nature. U-shaped MNF sensor: Reproduced under the terms of the CC BY license~\cite{xie2024optical}. Copyright 2024, The Authors, published by Wiley-VCH. Mechanoluminescent fiber: Reproduced with permission~\cite{luo2026self}. Copyright 2025, Wiley-VCH. Octopus-inspired arm: Reproduced under the terms of the CC BY 4.0 license~\cite{deldottore2026peripheral}. Copyright 2026, The Authors, published by Springer Nature.}
  }

\section{Optical Transduction Principles and Device Design}\label{sec:2}

Mechanical contact alters the geometry, strain or material state of a sensitive structure, producing a corresponding optical response. Reflective devices modulate optical signals through displacement or tilting of a reflective surface. Waveguides exhibit changes in transmission loss, scattering or coupling as light-guiding conditions change. Gratings and interferometric structures convert strain into wavelength shifts or changes in spectral shape, and mechanoluminescent materials emit light directly under mechanical stimulation. Different transduction mechanisms determine how optical signals correspond to contact states (Table~\ref{tab:1}). Even when optical intensity is used for readout, elastic reflective elements, continuous soft waveguides and luminescent composites provide different types of contact information.

Device design must produce measurable optical changes in response to the target contact quantity and establish an identifiable relationship between that quantity and the output signal. Springs, membranes and cladding determine how loads are transmitted to optically sensitive regions. Differential channels, color encoding and multiple resonance responses provide observables for distinguishing different stimuli. Owing to their small fiber dimensions, micro/nanofibers (MNFs) exhibit distinct local bending and optical mode-field responses and can be used in sensing structures based on propagation loss, localized field modulation or distributed reflection.

\subsection{Reflection, Focusing and Spot Displacement}\label{sec:2.1}

Reflective devices use the movement of an elastic layer or mirror to alter optical coupling between an emitter and a detector. Compression, translation or tilting induced by contact loads changes the intensity of reflected light received by the detector, which is converted into a photocurrent or voltage output. Early two-dimensional arrays stacked LEDs, photodetectors and pressure-sensitive layers, using local changes in optical intensity at each element to characterize the load distribution~\cite{nomura1985two}. Parallel-gripper taxels~\cite{cirillo2021tactile} and distributed interactive skins~\cite{cirillo2017distributed} adopted the same local readout approach, connecting contact signals to local acquisition circuits. These devices achieve spatial localization through the arrangement of sensing elements.

Besides the optoelectronic components, force-transmission mechanisms can adjust the relationship between load and optical output. A two-stage spring fingertip array arranged a flexible planar spring at the dome tip in series with a stiffer linear compression spring~\cite{konstantinova2017fingertip}. Under light contact, displacement was generated mainly by the planar spring, while increasing loads further compressed the linear spring, allowing both initial contact detection and measurement of larger loads within a limited travel range. The two-stage displacement moved a reflective surface, enabling readout through changes in returned optical intensity. In a prism-tipped optical-fiber array, light was directed toward a mirror, and the returned intensity was modulated through changes in the distance between the mirror and the fiber ends~\cite{ataollahi2010tactile}. The former adjusted the conversion from load to displacement, and the latter the conversion from displacement to optical intensity. Changes in spring stiffness, preload and initial spacing alter the calibration relationship between optical intensity and load~\cite{konstantinova2017fingertip,ataollahi2010tactile}. An occlusion-based optoelectronic pressure sensor placed pyramidal protrusions inside a silicone housing between an LED and a photodiode~\cite{li2023intentional}. The protrusions partially blocked the optical path and deformed with the housing under pressure, further reducing the received light. 

Focusing modulation senses contact through an interface with variable curvature. A liquid-membrane lens uses liquid to transmit loads, so that membrane curvature and the degree of beam focusing vary with load, and a photoresistor measures the resulting optical intensity (Figure~\ref{fig:2}a-i,ii)~\cite{hu2023flexible}. The liquid and membrane jointly determine the beam distribution under pressure, converting changes in membrane shape into an optical output. A device combining triboelectric and optical sensing further employed two signal branches~\cite{yang2024novel}. Its triboelectric contact membrane provided signals indicating contact and separation events, which the control circuit used to switch an externally powered light source on or off. Contact force was transmitted hydraulically to change the curvature of the liquid-membrane lens, after which a silicon photovoltaic cell provided quantitative readout. As long as contact was maintained, the control system kept the light source on, allowing the optical branch to measure the load continuously~\cite{yang2024novel}.

Multiple optoelectronic channels can encode contact direction through differences in their responses to reflective-surface movement. Under compression and shear, the soft pillars of PapillArray moved internal reflective structures, producing light spots through pinholes that induced different photocurrent changes in each quadrant~\cite{khamis2019novel}. Three-dimensional forces and displacements were then reconstructed through multivariable calibration (Figure~\ref{fig:2}a-iii,iv)~\cite{khamis2019novel}. A three-dimensional tactile sensor with a foldable optical path embedded in a rigid–soft coupled structure (FOCERS) instead used the LED itself as the contact point~\cite{li2025focers}. When the LED moved under load, the optical channel enclosed by its bottom surface and a square channel in the rigid support changed accordingly, allowing the light received by four surrounding photodiodes to indicate three-dimensional forces. The rigid–soft coupled structure enabled the device to withstand pressures exceeding 50 times its measurement range without damage~\cite{li2025focers}. Lateral displacement of an emitter relative to a detector is another way to encode shear direction. In one such device, a vertical-cavity surface-emitting laser (VCSEL) and a photodiode were separated by a 180 μm-thick silicone sensing layer~\cite{missinne2012two}. Shear displaced the VCSEL laterally, changing the optical coupling between them. When the VCSEL was aligned with the center of a circular photodiode, only shear magnitude could be measured. After alignment with the edge of a square photodiode, the output varied only with displacement perpendicular to that edge. Two VCSELs aligned with adjacent edges and illuminated in a time-division sequence therefore enabled separate readout in two directions~\cite{missinne2012two}. Multiple four-element modules were assembled into a distributed skin~\cite{cirillo2014artificial}. Each module comprised a 2×2 array of four infrared LED–phototransistor pairs covered by a silicone layer containing white reflective surfaces and black partition walls. Normal and tangential forces were estimated from combinations of the four voltage outputs. A 6×6-module prototype had a normal full-scale range of approximately 8 N, a tangential range of approximately ±2 N and an estimation accuracy of approximately 0.5 N~\cite{cirillo2014artificial}.

Monolithic gallium nitride (GaN) devices integrate light-emitting and receiving elements on the same substrate, which reduces the number of discrete optoelectronic components. Contact loads are transmitted through polydimethylsiloxane (PDMS) reflective domes or optical-fiber and mirror structures. Ref.~\cite{luo2025optical} employed a CaCO₃-doped PDMS reflective dome whose deformation under load modulated the signal received by an on-chip photodetector. Ref.~\cite{shi2026multi} embedded optical fibers in a flexible PDMS structure, where mirror tilting caused the photocurrents of quadrant receiving elements to vary with loading direction, providing information on contact direction. Both devices integrated emitting and receiving elements on-chip but differed in their mechanical load paths and directional sensing methods~\cite{shi2026multi,luo2025optical}. An array of organic diodes instead allowed the same device structure to function as either an emitter or a photodetector. Rubrene/C\textsubscript{60} diodes could switch between electroluminescence and photodetection modes~\cite{kielar2019organic}. During operation, the central diode emitted light, while surrounding diodes received light reflected from the inner reflective coating of an elastic dome that deformed under pressure. Changes in open-circuit voltage were then used to measure contact force~\cite{kielar2019organic}.

\subsection{Waveguide Loss, Scattering and Coupling}\label{sec:2.2}

Bending, stretching or local compression changes optical confinement and propagation paths in waveguides, altering the output optical intensity. Early microbending optical-fiber arrays used overlapping fiber layers to convert local pressure into periodic microbending, coupling higher-order guided modes into radiation modes~\cite{jenstrom1989fiber}. The resulting intensity attenuation was measured using an LED and a PIN photodiode. Stretchable waveguides in a soft hand employed a transparent polyurethane core and low-refractive-index silicone cladding~\cite{zhao2016optoelectronically}. Stretching increased the optical path length, and bending and local pressing altered the light-guiding geometry. Scattering at the core–cladding interface and absorption in the cladding jointly affected the output. Elastic waveguides coated with a metallic reflective layer measured pressure, strain or curvature through light leakage caused by increased microcracking in the coating during deformation~\cite{to2015highly,to2018soft}. All these approaches characterize deformation through changes in optical intensity, but the modulation mechanism differs: microbending-induced mode coupling in the fiber arrays, changes in path and geometry in the soft waveguides, and cracking of the coating in the metal-coated waveguides.

Planar waveguides allow optical modulation structures to be integrated into thin-film skins, and different pressure-sensitive interfaces provide distinct modulation mechanisms. A transparent pressure-sensitive skin generated photocurrent responses through changes in guided modes and light escape caused by PDMS compression~\cite{ramuz2012transparent}. In a dynamic tactile array, the output intensity of individual waveguides was modulated through interfacial scattering when the contact layer touched an exposed core~\cite{yun2014polymer}. The former relied primarily on deformation of the light-guiding film itself; the latter used the pressure-dependent contact interface as the main modulation region. The refractive index and mechanical properties of the contact layer affect optical coupling, load transmission and recovery after unloading.

Device design can be adjusted through the interface by introducing graphene, through the material by selecting the polymer contact layer, and through the waveguide cross-section by using the Mosquito fabrication method. In graphene waveguides, applied pressure changed the effective contact area among the waveguide core, graphene and PDMS, modulating the absorption of guided light by graphene~\cite{kim2018graphene}. Different polymer contact layers affected sensing responses through the mechanical properties of the material, and the Mosquito fabrication method through the waveguide cross-sectional structure~\cite{yin2025polymer}. Comparisons of contact-layer materials under identical test conditions showed that material selection influenced both sensitivity and dynamic recovery characteristics~\cite{yun2014polymer}.

Tactile coverage can be extended either by moving the sensing location or by obtaining spatial observations from multiple overlapping optical paths within a continuous medium. A flexible waveguide film measured local deformation through a single transmission path~\cite{zhang2025flexible}. After installation on a curved surface, the robot pressed successive locations and combined the measurements with known sensing positions to generate a stiffness distribution. In a transparent distributed skin, 8 LEDs and 8 phototransistors were arranged around the perimeter~\cite{levi2013soft}. The LEDs were illuminated sequentially and the detectors read out, producing 64 emitter–receiver pairwise observations per cycle. Local contact altered the optical intensity passing through the affected regions through frustrated total internal reflection and waveguide deformation. The magnitude of these changes across channels depended on contact location, area, pressure and optical-path coverage. Responses from the overlapping channels were used to reconstruct the contact distribution~\cite{levi2013soft}. A finger-pad skin for a granular-jamming gripper stacked three transverse and three longitudinal soft waveguides with glycerol liquid cores into a 3×3 array~\cite{mo2025empowering}. Six channels of 940 nm infrared optoelectronic output were classified using a random forest to distinguish 49 classes of contact locations comprising intersections, regions and line segments, achieving an accuracy of 99.80\% under ten-fold cross-validation~\cite{mo2025empowering}.

Contact coupling can also be used to construct three-dimensional lightguide sensing networks. Optical Lace threaded active and passive soft lightguides through a flexible lattice~\cite{xu2019optical}. Under compression, the contact length between them increased, strengthening the optical intensity coupled into the passive outputs. Intensity ratios between neighboring outputs were used to estimate contact locations. The network was used in a touch-sensitive musical instrument, where pressing locations selected notes and changes in optical intensity controlled volume. In a variable-stiffness cylindrical lattice, local forces estimated from individual channels were combined with a lattice compression model to reconstruct deformation in different sections. The stretchable lightguide for multimodal sensing (SLIMS) employed dyed cores, transparent cores and discrete dyed regions, using red–green–blue (RGB) channels to measure color and intensity~\cite{bai2020stretchable}. In a soft glove, one dual-core lightguide was arranged along each finger, with three dyed regions corresponding to three joints. Multijoint bending was reconstructed by combining single-joint calibration with responses from the dyed core. Changes in intensity from the transparent core were used to identify external pressing, while hue indicated the pressing location and qualitatively reflected the magnitude of the applied force (Figure~\ref{fig:2}b)~\cite{bai2020stretchable}.

Scattering media can also convert mechanical deformation into changes in optical intensity. Compression of a flexible porous monolith altered its internal scattering paths, and the response was measured using a commercial infrared reflector~\cite{hayase2023optical}. The sensitive optical paths were formed by the internal structure of the porous material, allowing mechanical stimuli to be detected through a small number of intensity channels. Soft waveguides, lightguide networks and porous media modulate optical intensity through bulk deformation, contact coupling and internal structural compression, respectively. An octopus-inspired flexible optical tactile sensor similarly exploited changes in scattering within a porous medium~\cite{li2026octopus}. Compression of a porous shape-memory polymer film reduced its porosity and the number of air–polymer interfaces, weakening scattering and consequently increasing the reflected light received by a photodetector inside the cavity. Beside the force-sensing cavity, a color module illuminated objects with a white-light LED and measured reflected light using an RGB color sensor. The two optical paths were isolated from each other, preventing crosstalk between the force and color signals~\cite{li2026octopus}.

\subsection{Micro/Nanofibers and Localized Sensitive Structures}\label{sec:2.3}

Part of the guided optical field in micro/nanofibers extends beyond the fiber surface, making them sensitive to changes in the surrounding medium and fiber bending. Flexible encapsulation converts surface contact into local fiber deformation, modulating optical signals through bending loss, mode changes or evanescent-field interactions. Fiber dimensions affect the optical mode field and bending response, while fiber embedding depth, encapsulation thickness and contact location determine how loads are transmitted to the fiber. Micro/nanofiber hardness probes~\cite{tang2021optical} characterize hardness through differences in responses induced by contact deformation of different soft materials, while multi-axis force devices use asymmetric structures to produce distinguishable changes in optical intensity under loads applied in different directions~\cite{xie2024optical}. An S-shaped micro/nanofiber chip inspired by the interlocking structure of the epidermis and dermis encapsulated an optical fiber in a PDMS film~\cite{yang2025bionic}. Two elastic resins of different stiffnesses formed conical interlocking microstructures that amplified local stresses in the contact region. The device produced wavelength shifts in two intermodal interference resonance dips, which were used to measure microcontact forces and temperature and distinguish material hardness~\cite{yang2025bionic}.

The response of a U-shaped micro/nanofiber depends strongly on loading geometry, with both the normal-force response slope and effective measurement range varying with loading position (Figure~\ref{fig:2}c-i,ii; numerical values are given in Section~\ref{sec:5.1})~\cite{xie2024optical}. The authors used a laser light source and a PD10A-200M-4ch multichannel photodetector and estimated a force resolution of approximately 120 μN from the detector noise floor~\cite{xie2024optical}. This estimate was based on a fixed loading position and readout configuration. The experiments also limited the normal loading rate to within 5.9 N/s and required the sensitive region to be mounted on a substrate unlikely to introduce additional stretching or bending. Used as a taxel, the device would therefore need calibration that includes contact location, encapsulation deformation, loading rate and readout noise~\cite{xie2024optical}.

Fiber topology can exploit geometric asymmetry to produce localized responses that vary with loading direction. Polymer knot-shaped optical-fiber (OFN) sensors~\cite{pan2023knot} encode pressure, friction and shear through anisotropic microbending, while S-shaped tapered devices~\cite{gao2026s} identify sliding direction through dynamic changes in optical intensity. Both structures modulate optical paths differently under normal preload and tangential motion, acquiring contact information from temporal variations in optical intensity. The directional response of an OFN sensor arises from its topology and microbending effects. In a dual-arm manipulation demonstration, a multimodule OFN device combined pressure, hardness and texture sensing~\cite{yu2023optum}. A fingerprint-inspired flexible optical sensor used elastic resin and PDMS of different stiffnesses to mimic the epidermal and dermal layers and encapsulated a microfiber within the soft layer~\cite{jiang2021finger}. Parallel ridges on the top surface enhanced friction and vibration during sliding. The transmitted optical intensity from the same optical path reflected both normal forces over a range of 0–20 N and contact slip~\cite{jiang2021finger}.

Localized optical fields can also contribute to tactile transduction through material resonance. A flexible plasmonic optical fiber exploited the localized surface plasmon resonance (LSPR) of gold nanoparticles (GNPs) to convert mechanical deformation into optical responses, allowing tactile signals to be read through transmitted optical intensity at a fixed wavelength (Figure~\ref{fig:2}c-iii,iv)~\cite{guo2023flexible}. The local mode field of the micro/nanofiber and the resonance characteristics of the gold nanoparticles provide distinct optical sensing mechanisms, while flexible encapsulation transmits contact loads to the sensitive region~\cite{guo2023flexible}.

\subsection{Fiber Gratings and Distributed Sensing}\label{sec:2.4}

A fiber Bragg grating (FBG) senses strain through shifts in its reflection peak wavelength, with the central wavelength given by Equation~\eqref{eq:1}:
\begin{equation}
\lambda_{B}=2n_{\mathrm{eff}}\Lambda
\label{eq:1}
\end{equation}
where $n_{\mathrm{eff}}$ is the effective refractive index and $\Lambda$ is the grating period. Contact loads are transmitted to the optical fiber through membranes, bridge-type force-transmission mechanisms or flexible cladding, inducing strain and shifts in the reflection peak wavelength. The reflection peak wavelength can serve as a spectral identifier for the sensing location, allowing multiple gratings along a single fiber to be read through wavelength-division multiplexing and thereby extending local load measurement to an array. Early studies of grating-based tactile sensing~\cite{heo2006tactile} and distributed optical-fiber sensing~\cite{heo2008tactile} demonstrated such structures, with the FBG approach acquiring contact information from shifts in the reflection peaks of predefined gratings.

Grating layout and force-transmission structures determine whether an array uses local measurements or overlapping responses. A helical flexible mechanism concentrated loads onto individual FBG elements, enabling point-by-point readout through mechanical strain conversion~\cite{li2018high}. Polymer optical-fiber arrays~\cite{leong2024artificial} and deeply embedded optical-fiber skins~\cite{massari2022functional} instead transmitted strain through continuous encapsulation, causing a single contact to produce responses in multiple gratings. Local force transmission establishes a relatively direct correspondence between sensing elements and contact regions. Continuous cladding produces different combinations of multichannel responses at different contact locations. The latter approach can estimate contact locations between grating elements from relative changes across channels~\cite{massari2022functional,leong2024artificial}.

Artificial intelligence-assisted FBG skins~\cite{li2023artificial}, Fibertouch~\cite{qiao2025fibertouch} and biomimetic robotic fingers~\cite{mao2024bioinspired} use multichannel wavelength or spectral responses to estimate contact force, location or tactile categories from reflected signals. A composite fingertip sensor suspended four FBG-containing optical fibers within a 3D-printed metal elastomer to measure three-axis fingertip forces, achieving sensitivities of 225, 229 and 97 pm/N in the x, y and z directions within ±10 N~\cite{li2025fingertip}. Five additional FBGs were embedded in a PDMS film on the finger pad, and a random forest and a backpropagation (BP) neural network were used for classification and regression of single-point and two-point pressing. The classification accuracy was 99\%~\cite{li2025fingertip}.

Denser along-fiber arrangements can distinguish sensing elements by spatial position rather than by different central wavelengths. A double-layer staggered weak-FBG skin employed weak gratings with identical central wavelengths and used optical frequency-domain reflectometry (OFDR) to distinguish reflected signals from different positions along the fiber and detect contact-induced wavelength shifts (Figure~\ref{fig:2}d-i,ii)~\cite{fu2026double}. Along-fiber demodulation allows a single optical fiber to support both multipoint sensing and multiplexed transmission~\cite{fu2026double}.

A distributed micro/nanofiber skin used OFDR to measure Rayleigh backscatter (RBS) and its correlation spectra, with a learning model estimating contact locations and forces (Figure~\ref{fig:2}d-iii,iv)~\cite{li2026continuous}. Unlike the predefined discrete reflective sections of weak FBGs~\cite{fu2026double}, continuous Rayleigh backscatter can be used to extract local responses along the fiber. Spatial information originates differently in the two directions. The x direction along the fiber can be resolved through OFDR frequency-to-position mapping; the y direction perpendicular to the fiber depends on strain propagation through the encapsulation and combinations of responses from multiple fibers. The 9.88 μm reported in the original study represents the OFDR spatial resolution along the fiber, and a 150-point correlation window corresponds to a sensing spatial resolution of approximately 1.48 mm~\cite{li2026continuous}. Neither quantity is equivalent to the two-dimensional contact localization error of the learning model.

\subsection{Interference and Resonance Modulation}\label{sec:2.5}

Interferometric tactile sensing uses contact-induced changes in optical path difference, cavity length or mode coupling to modulate interference signals. A Mach–Zehnder interferometer (MZI) can be constructed from optical-fiber sections with different core diameters~\cite{li2024learning}. Flexible cladding converts contact loads into fiber strain and bending, thereby changing the intermodal interference spectrum~\cite{xu2022highly}. In a Fabry–Pérot interferometer (FPI), the optical cavity can be formed between a fiber end face and a pressure-sensitive diaphragm~\cite{keser2021fiber}. Under the corresponding optical-path conditions, the round-trip phase of light within the cavity is expressed by Equation~\eqref{eq:2}:
\begin{equation}
\varphi=\frac{4\pi nL}{\lambda}
\label{eq:2}
\end{equation}
where $n$ is the refractive index within the cavity, $L$ is the cavity length and $\lambda$ is the vacuum wavelength. Contact-induced changes in cavity length can be measured through spectral changes or optical intensity responses at a fixed wavelength. An FPI surface-roughness probe employed a metallic hemispherical contact tip and fixed-wavelength optical readout, converting the dynamic responses generated during scanning of surfaces of different roughness into time series of optoelectronic signals~\cite{keser2021fiber}. Discrete wavelet transform and machine learning algorithms were then used to identify predefined roughness categories. An MZI tactile pad instead used full-spectrum features to identify interference spectral shapes corresponding to different loading locations and force levels (Figure~\ref{fig:2}e-i,ii)~\cite{li2024learning}. Interferometric tactile sensing can be used both to detect dynamic contact events and to identify spatial contact states. An early miniature design used optical feedback from an external mirror into a surface-emitting laser, so that one chip served as light source and interferometer and no lenses or optical isolator were needed~\cite{yamamoto1995optical}. More recently, an MZI formed by a tapered mode-field-mismatch structure and embedded in PDMS reached a force sensitivity of 23.636 nm/N with a detection limit of 0.746 mN~\cite{li2024ultrasensitive}. Filtering weak high-order modes with fast and inverse fast Fourier transforms reduced its repeatability error below 1\%~\cite{li2024ultrasensitive}. In a sensor for robotic bronchoscopy, four FPIs with ellipsoidal air cavities in a hollow spiral elastomer resolved 0.57, 2.46 and 4.92 mN along the x, y and z axes within $\pm$3 N, and a thermo-mechanical coupling model compensated temperature-induced baseline drift~\cite{wang2026fiber}.

Full-spectrum readout preserves multiple deformation-sensitive spectral features, allowing the overall spectral shape to identify predefined loading locations, force levels, object shapes and Braille inputs. Recognition performance in a single-MZI tactile pad~\cite{li2024learning} and a ring-shaped soft sensor~\cite{li2025integrating} was affected by environmental and training conditions. Ref.~\cite{li2024learning} used 25 °C as the initial condition and, after incorporating data from the corresponding temperature range, maintained relatively good recognition performance for known shapes and angles within 20.4–27 °C. The main loading experiments in Ref.~\cite{li2025integrating} were conducted at 24 °C, and the temperature-perturbation tests covered four temperature intervals within 19.7–25.3 °C. Both studies observed that recognition performance with a fixed-temperature model declined as the temperature difference increased and improved after data from the tested temperature ranges were incorporated.

Multiparameter decoupling exploits differences in optical responses to different target quantities, while demodulation converts these optical changes into signals suitable for computation. A macrobend FBG device combined grating resonance with whispering-gallery mode (WGM) resonance~\cite{shang2023soft}. The two types of resonance exhibited different force-response slopes and opposite temperature-response directions, allowing force and temperature to be recovered separately through a calibrated sensitivity matrix (Figure~\ref{fig:2}e-iii,iv). Dual-comb technology was used for rapid acquisition of resonance spectra, whereas force–temperature decoupling relied on the differential sensitivities of the two resonances and matrix inversion. In a flexible optical-fiber ring resonator, Pound–Drever–Hall (PDH) demodulation was used and the error signal was divided into two branches~\cite{wang2025optical}. One branch provided feedback through a proportional–integral (PI) servo controller to adjust the laser frequency and track the cavity resonance. The other was used to analyze contact-induced dynamic responses and, together with a learning model, enabled Braille recognition. PDH demodulation stabilizes resonance readout and improves sensitivity but does not directly decouple temperature from contact force~\cite{wang2025optical}.

\subsection{Direct Emission through Mechanoluminescence}\label{sec:2.6}

Mechanoluminescence (ML) generates optical signals under mechanical stimulation. An optical-fiber tactile device based on a ZnS:Cu and PDMS composite operated without external optical excitation, transmitted contact deformation to the luminescent material, collected the resulting light through an optical fiber and detected contact and slip events through photon counting (Figure~\ref{fig:2}f-i,ii)~\cite{luo2026self}. The sensing end thus achieved self-powered optical transduction. This self-powered optical fiber tactile sensor (SOFTS) combined mechanical excitation with optical-fiber transmission and used temporal information from luminescence peaks to trigger robotic actions~\cite{luo2026self}. Regarding the luminescence mechanism of elastomeric composites such as ZnS:Cu/PDMS, one study confirmed that their emission arises from ZnS:Cu electroluminescence induced by triboelectrification at the particle–PDMS interface~\cite{lee2024interfacial}. Luminescence from centrosymmetric, non-piezoelectric Lu\textsubscript{3}Al\textsubscript{5}O\textsubscript{12}:Ce\textsuperscript{3+}/PDMS elastomers was also shown to originate from contact electrification~\cite{wang2022contact}. A greater difference in the triboelectric series positions of the two phases produces stronger luminescence, and selecting a suitable inorganic–organic combination can increase luminescence intensity by approximately 20 times (Figure~\ref{fig:2}f-iii,iv)~\cite{pan2024quantifying}. Maintaining brightness under large tensile deformation also requires a highly elastic matrix with strong negative triboelectric properties~\cite{jeong2025super}.

Unlike devices that continuously probe structural states using an external light source, mechanoluminescence generates optical signals during mechanical deformation, with emission peaks influenced by both deformation magnitude and rate. In the robotic grasping demonstration of this device, no further luminescence was generated during the stable holding phase after the load reached equilibrium~\cite{luo2026self}. Dynamic processes such as contact initiation and slip can therefore be identified from luminescence peak events. Continuous measurement of absolute load requires an established relationship between the load during the holding phase and an observable signal. Although the transduction element requires no external optical excitation, photon counting, signal processing and robotic actuation still require power~\cite{luo2026self}. The dependence of luminescence on loading speed can provide an additional observable. A LiTaO\textsubscript{3}:Tb\textsuperscript{3+} film produced distinguishable green luminescence at a strain of only 500 με and used a custom parameter Q to characterize the rate of stress change~\cite{yang2024microstrain}. Luminescence under small loads may fall below the detection threshold. In a composite device incorporating ZnS:Cu/PDMS and piezoelectric silk fibroin/PDMS, a photomultiplier tube detected no luminescence at 1 N, with reliable detection beginning at 10 N~\cite{kim2024mechanoluminescent}. Recognition of handwritten letters relied solely on the piezoelectric voltage~\cite{kim2024mechanoluminescent}. When a mechanoluminescent layer serves as the spacer~\cite{zhang2022self} or intermediate layer~\cite{lin2024self} of a contact–separation triboelectric nanogenerator, the electrical signal responds primarily to pressing, whereas the optical signal responds to tensile or bending strain. Combinations of these responses can be used to distinguish pressing, bending and stretching~\cite{zhang2022self}.

\adrrfig{fig:2}{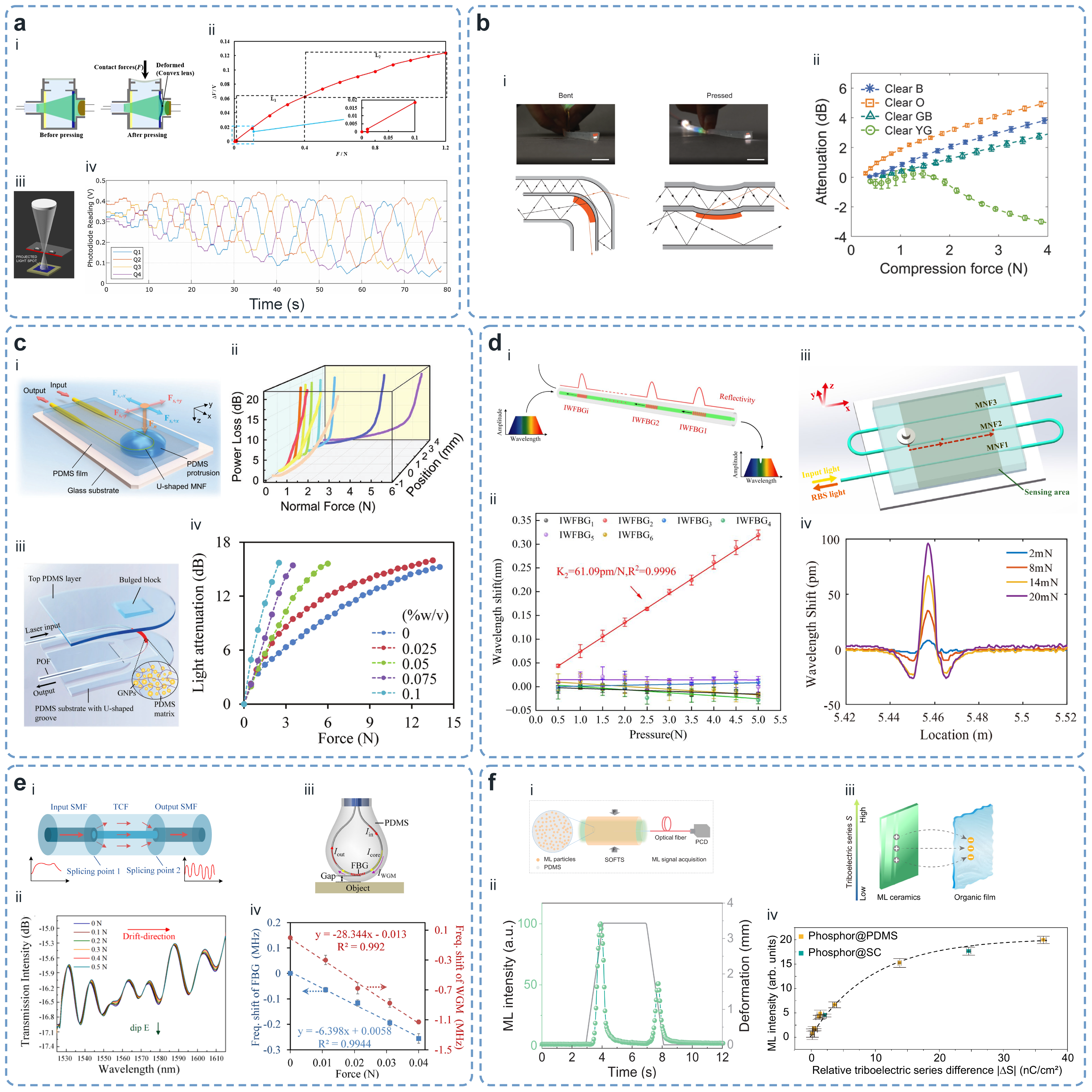}%
  {Optical transduction mechanisms in non-imaging optical tactile sensors.}%
  {%
\capline{a) Reflection, focusing and spot displacement: i) liquid-membrane lens before and after pressing; ii) output-voltage change versus contact force. Reproduced with permission~\cite{hu2023flexible}. Copyright 2023, Optica Publishing Group. iii) PapillArray pillar projecting a light spot onto a quadrant photodiode; iv) quadrant-photodiode outputs for a spiral stimulus (time-axis title added from the original figure). Reproduced with permission~\cite{khamis2019novel}. Copyright 2019, Elsevier.}
\capline{b) Propagation loss, scattering and coupling: i) stretchable dual-core lightguide (SLIMS) when bent and pressed (scale bars, 1 cm); ii) attenuation in four color channels versus compression force. Reproduced with permission~\cite{bai2020stretchable}. Copyright 2020, The American Association for the Advancement of Science.}
\capline{c) Local field and material-sensitive structures: i) U-shaped micro/nanofiber (MNF) under a PDMS protrusion; ii) power loss versus normal force at different protrusion positions. Reproduced under the terms of the CC BY license~\cite{xie2024optical}. Copyright 2024, The Authors, published by Wiley-VCH. iii) PDMS optical fiber doped with gold nanoparticles (GNPs); iv) attenuation of transmitted 532 nm light versus pressing force. Reproduced with permission~\cite{guo2023flexible}. Copyright 2023, Wiley-VCH.}
\capline{d) Bragg reflection and along-fiber Rayleigh scattering read by optical frequency-domain reflectometry: i) identical-wavelength weak fiber Bragg gratings (IWFBGs); ii) wavelength shifts of IWFBG1–6 under pressure on IWFBG2. Reproduced under the terms of the Optica Open Access Publishing Agreement~\cite{fu2026double}. Copyright 2026, Optica Publishing Group. iii) skin with three MNFs in series read by Rayleigh backscatter; iv) wavelength shift versus location for loads of 2–20 mN. Reproduced with permission~\cite{li2026continuous}. Copyright 2026, Wiley-VCH.}
\capline{e) Interference and resonance: i) single-mode–thin-core–single-mode fiber Mach–Zehnder interferometer (MZI) tactile pad; ii) transmission spectra under 0–0.5 N. Reproduced with permission~\cite{li2024learning}. Copyright 2024, Elsevier. iii) macrobent FBG tip in PDMS forming whispering-gallery modes (WGMs); iv) FBG and WGM resonance shifts versus force. Reproduced with permission~\cite{shang2023soft}. Copyright 2023, American Chemical Society.}
\capline{f) Mechanoluminescence (ML): i) self-powered optical fiber tactile sensor (SOFTS) read by photon counting; ii) ML intensity and deformation over a compression–release cycle. Reproduced with permission~\cite{luo2026self}. Copyright 2025, Wiley-VCH. iii) electron transfer at an inorganic–organic ML interface; iv) ML intensity versus relative triboelectric series difference. Reproduced under the terms of the CC BY 4.0 license~\cite{pan2024quantifying}. Copyright 2024, The Authors, published by Springer Nature.}%
  }

\begin{table*}[!tbp]
\caption{Main optical modulation processes and representative devices}\label{tab:1}
\begin{tabular*}{\textwidth}{@{}>{\raggedright\arraybackslash}p{0.176\dimexpr\textwidth-8\tabcolsep\relax}>{\raggedright\arraybackslash}p{0.212\dimexpr\textwidth-8\tabcolsep\relax}>{\raggedright\arraybackslash}p{0.176\dimexpr\textwidth-8\tabcolsep\relax}>{\raggedright\arraybackslash}p{0.312\dimexpr\textwidth-8\tabcolsep\relax}>{\raggedright\arraybackslash}p{0.124\dimexpr\textwidth-8\tabcolsep\relax}@{}}
\toprule
\textbf{Main process} & \textbf{Representative device configurations} & \textbf{Measured signals} & \textbf{Tactile functions and characteristics} & \textbf{Refs.} \\
\midrule
Reflection, focusing and spot displacement & Elastic reflective elements, liquid-membrane lenses, quadrant-sensing soft pillars, integrated GaN units & Photocurrent, voltage, quadrant differential signals & Local loads and directions; force-transmission structures determine displacement, while optical paths and channel relationships determine outputs & \cite{khamis2019novel,shi2026multi,hu2023flexible} \\
Propagation loss, scattering and coupling & Soft waveguides, color-encoded elastic lightguides, optical-fiber knots, three-dimensional lightguide networks & Transmitted intensity, relative intensity, multiband outputs & Strain, curvature, contact and friction; optical-path layout and coupling relationships provide spatial information & \cite{zhao2016optoelectronically,bai2020stretchable,xu2019optical,pan2023knot} \\
Local mode fields and plasmonic resonance & Micro/nanofibers, flexible plasmonic optical fibers & Transmitted intensity & Local contact, hardness discrimination and multi-axis sensing; responses depend on fiber dimensions and flexible encapsulation & \cite{tang2021optical,xie2024optical,guo2023flexible} \\
Bragg reflection and along-fiber backscatter & FBG arrays, weakly reflecting gratings, distributed micro/nanofibers & Reflection wavelengths, along-fiber correlation spectra & Multipoint coverage, force and location estimation; layout, strain propagation and multiplexed demodulation jointly provide spatial observations & \cite{massari2022functional,fu2026double,li2026continuous} \\
Interference and resonance & MZI tactile pads, FPI diaphragms, FBG/WGM, optical-fiber rings & Spectral features or fixed-wavelength intensity & Force and location recognition, temporal texture signals, multiparameter separation; spectral features and readout methods selected according to target variables & \cite{shang2023soft,keser2021fiber,li2024learning} \\
Mechanoluminescence & ML composites and light-collecting optical fibers & Photon counts, temporal profiles of light peaks & Contact and slip events; direct light emission under mechanical stimulation, with outputs dependent on deformation magnitude and rate & \cite{luo2026self} \\
\bottomrule
\end{tabular*}
\begin{tablenotes}
Note: Entries are organized by the main optical process; a single device may exploit multiple processes, and its actual readout is described according to the specific experiments.
\end{tablenotes}
\end{table*}

\section{Structural and System Integration}\label{sec:3}

Optical transduction mechanisms determine how contact is converted into measurable signals, while system integration influences signal acquisition and its use in robotics. Sensitive structures can be deployed at fingertips, on flexible surface layers or throughout deformable bodies. Light sources and detectors can be integrated near the sensing region or connected remotely through optical fibers. This section first summarizes materials and fabrication routes for optical sensing structures and then compares system architectures in terms of sensor deployment, spatial coverage, signal multiplexing and structural encapsulation. These architectures determine the system-level constraints on spatial coverage and feedback timing, the second of the three considerations set out in the Introduction.

\subsection{Materials and Fabrication}\label{sec:materials}

Soft optical tactile structures need materials that guide light while deforming with the contact layer, and fabrication methods that place the optical path where loads act. Fabrication of fiber-optic tactile sensors has been reviewed in detail~\cite{zhu2025advancements,zhang2024optical}; the examples below follow the five groups in Figure~\ref{fig:materials}. Step-index elastomer waveguides combine a high-index core with a lower-index cladding. Elastic polyurethane fibers (refractive index 1.52) dip-coated with a PDMS cladding (1.417) were coupled to a micro-LED and a photodiode and reached a fracture strain of 438\%~\cite{fan2026intelligent}. Commercial PDMS can be replicated into stretchable waveguides, and integrating VCSELs and photodiodes with them allowed losses to be measured under bending and 30\% elongation~\cite{missinne2014stretchable}. Melt coextrusion of two thermoplastic elastomers produced step-index fibers at several hundred meters per hour that guided light reversibly at strains of up to 300\%~\cite{leber2019stretchable}. Hydrogels extend the strain range further: alginate–polyacrylamide core/cladding fibers sustained 700\% axial strain~\cite{guo2016highly}, and a double-network hydrogel waveguide was stretched to 600\% (Figure~\ref{fig:materials}a)~\cite{li2025nerve}. Continuous coaxial spinning with in situ photopolymerization produced refractive-index-matched core–cladding hydrogel fibers with losses of 0.314–0.374 dB/cm in air for diameters of 600–1200 μm~\cite{zhou2026coaxial}. Self-healing chemistry adds damage tolerance: a transparent poly(urethane-urea) elastomer combining hydrogen bonds with room-temperature exchange of aromatic disulfide bonds healed cuts in lightguides without external intervention~\cite{bai2022autonomous}.

Micro/nanofibers are drawn from standard optical fibers. A two-step drawing process yielded free-standing silica wires as thin as 50 nm with atomic-level surface smoothness and an optical loss below 0.1 dB/mm~\cite{tong2003subwavelength}. Adiabatic stretching keeps the original fiber dimensions at both ends of the taper so that it can be spliced to standard fibers~\cite{brambilla2009optical}. For tactile use, the fragile taper is embedded in an elastomer. Glass micro/nanofibers embedded in thin PDMS layers sensed pressure through the transition from guided to radiation modes, with a sensitivity of 1870 kPa$^{-1}$, a detection limit of 7 mPa and a response time of 10 μs~\cite{zhang2020ultrasensitive}. A tapered polymer optical fiber was embedded in an S shape by two-step PDMS casting in a 3D-printed mold~\cite{gao2026s}, and in nanofiber skins the thickness of the PDMS film changed the response to pressing force (Figure~\ref{fig:materials}b)~\cite{tang2023optical}.

Fiber Bragg gratings are inscribed in the fiber before it is embedded. Shape deposition manufacturing embedded several FBGs in a polymer finger structure to measure contact and grasping forces~\cite{park2007force}. In a Ruffini-type skin, the grating fiber was encapsulated by two-step silicone casting in 3D-printed molds, and increasing the encapsulation thickness from 4 to 12 mm reduced sensitivity and enlarged the receptive fields (Figure~\ref{fig:materials}c)~\cite{massari2022functional}. Polymer optical fibers have a much lower Young's modulus than silica (about 3.3 GPa for poly(methyl methacrylate) versus 73 GPa), which raises their response to axial force. Their gratings, however, must be inscribed without damaging the fiber surface, and the polymer absorbs about 2\% water, which adds humidity sensitivity~\cite{webb2015fibre}. For instruments that must be sterilized, hydrofluoric-acid etching produced a step-reduced FBG with twin reflection peaks inside an enclosed package 4.5 mm in diameter, which remained stable after sterilization at 180 °C~\cite{li2025high}.

Mechanoluminescent layers are usually composites of phosphor microparticles in an elastomer. Mechanoluminescence can arise from fracture, friction or elastic deformation~\cite{zhuang2021mechanoluminescence}, and Cu- or Mn-doped ZnS emits under stress without pre-irradiation; in one formulation the phosphor was mixed into PDMS at a phosphor-to-PDMS weight ratio of 7:3~\cite{jeong2014bright}. Such composites can be applied to fibers: a stretchable fiber with a silicone core (refractive index 1.53) was coated segment by segment with ZnS:Cu, ZnS:Mn or mixed phosphor layers whose emission colors encoded the strain location~\cite{zhou2023distributed}. Printing is another route: shear-thinning ZnS/silicone inks were direct-ink-written into auxetic films, which were then encapsulated in silicone to even out the emission~\cite{sun2025printing}. Co-doping can add a second channel; a CaZnOS:Nd,Er fiber combined near-infrared mechanoluminescence with 980 nm-excited upconversion for temperature readout (Figure~\ref{fig:materials}d)~\cite{zheng2025flexible}.

Integrated forming reduces assembly steps. In a granular-jamming gripper skin, stretchable waveguides were molded in one piece around 1.5 mm iron-wire cores; glycerol was then injected as the liquid core, infrared emitters and receivers were inserted at the ends, and the cladding hardness changed the pressure response (Figure~\ref{fig:materials}e)~\cite{mo2025empowering}. Optical Lace used a 3D-printed elastomeric lattice as the load-bearing structure for its lightguides~\cite{xu2019optical}. Additive manufacturing can form the optics themselves. Soft thermoplastic waveguides printed with more than 100\% elastic strain showed transparency above 98\% and a loss below 0.22 dB/cm~\cite{udofia2020printed}, and a lens printed by digital light processing inside a Y-shaped waveguide redistributed light between two branches when deformed. The lens raised sensitivity more than 65-fold relative to the lens-free design~\cite{cafiso2026printed}. Stereolithography printing reduced an intensity-modulated surgical force sensor to 14.0 × 1.8 × 4 mm~\cite{bandari2020miniaturized}. For thin planar devices, continuative multistep photolithography produced a polymer-waveguide tactile layer less than 200 μm thick with a transmittance of up to 90\%~\cite{yun2020soft}, following earlier lithographic waveguide arrays~\cite{yun2014polymer}.

\adrrfig{fig:materials}{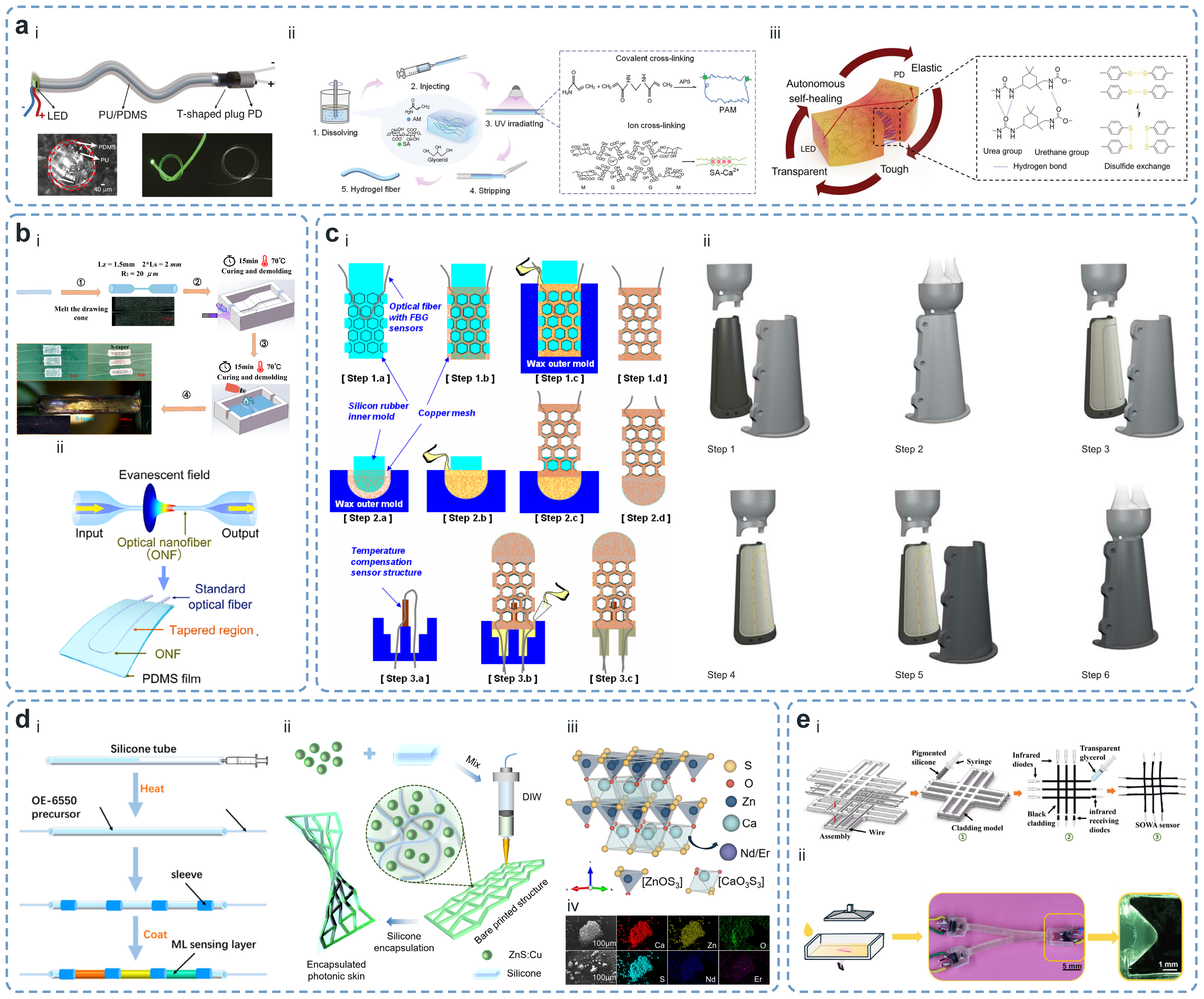}%
  {Materials and fabrication of non-imaging optical tactile sensing structures.}%
  {%
\capline{a) Waveguide and fiber materials: i) polyurethane-core, PDMS-clad elastic waveguide, its cross-section and knotted samples. Reproduced under the terms of the CC BY 4.0 license~\cite{fan2026intelligent}. Copyright 2026, The Authors, published by Springer Nature. ii) hydrogel optical fibers injected into a silicone tube and UV-cured, with covalent (polyacrylamide) and ionic (alginate–Ca\textsuperscript{2+}) cross-linking. Reproduced under the terms of the CC BY license~\cite{li2025nerve}. Copyright 2024, The Authors, published by Wiley-VCH. iii) self-healing poly(urethane-urea) waveguide based on hydrogen bonds and disulfide exchange. Reproduced under the terms of the CC BY-NC 4.0 license~\cite{bai2022autonomous}. Copyright 2022, The Authors, published by the American Association for the Advancement of Science.}
\capline{b) Micro/nanofiber drawing and packaging: i) polymer optical fiber tapered by heat drawing and cast in PDMS in a 3D-printed S-shaped groove. Reproduced under the terms of the Optica Open Access Publishing Agreement~\cite{gao2026s}. Copyright 2026, Optica Publishing Group. ii) optical nanofiber and a U-shaped nanofiber embedded in a PDMS film. Reproduced under the terms of the CC BY license~\cite{tang2023optical}. Copyright 2022, The Authors, published by Wiley-VCH.}
\capline{c) Gratings embedded in polymer: i) shape deposition manufacturing of a robot finger: FBG fibers on a silicone inner mold, wrapped in copper mesh and cast in polyurethane in wax molds. Reproduced with permission~\cite{park2007force}. Copyright 2007, IEEE. ii) two-step silicone casting in 3D-printed molds with the FBG fiber between the layers. Reproduced under the terms of the CC BY 4.0 license~\cite{massari2022functional}. Copyright 2022, The Authors, published by Springer Nature.}
\capline{d) Mechanoluminescent materials and composites: i) stretchable fiber cured in a silicone tube, divided into segments by sleeves and coated with phosphor/PDMS layers. Reproduced under the terms of the CC BY license~\cite{zhou2023distributed}. Copyright 2023, The Authors, published by Wiley-VCH. ii) direct ink writing of a ZnS:Cu/silicone ink into an auxetic structure, then silicone encapsulation, shown as a printing example only. Reproduced under the terms of the CC BY-NC license~\cite{sun2025printing}. Copyright 2025, The Authors, published by Wiley-VCH. iii) crystal structure of CaZnOS:Nd\textsuperscript{3+},Er\textsuperscript{3+}; iv) SEM images and elemental maps of the phosphor. Reproduced under the terms of the CC BY license~\cite{zheng2025flexible}. Copyright 2025, The Authors, published by Wiley-VCH.}
\capline{e) Integrated molding and 3D printing: i) one-piece molding of a 3 × 3 waveguide array around iron wires, followed by a glycerol core and infrared diodes. Reproduced under the terms of the CC BY license~\cite{mo2025empowering}. Copyright 2024, The Authors, published by Wiley-VCH. ii) soft acrylate polyurethane printed by digital light processing into a Y-shaped waveguide sensor, with a microscope image of its printed lens. Reproduced under the terms of the CC BY-NC license~\cite{cafiso2026printed}. Copyright 2026, The Authors, published by Elsevier.}%
  }

\subsection{Local Optoelectronic and Remote Readout}\label{sec:3.1}

Local optoelectronic units integrate light emitters, detectors and elastic force-transmission structures with amplification, analog-to-digital conversion and communication circuits. Modular reflective platforms use repeated units for deployment on robotic surfaces, and GaN devices integrate emitting and receiving elements on a single chip (Figure~\ref{fig:3}a). The former supports module rearrangement and replacement; the latter achieves monolithic integration of emitting and receiving functions. Both approaches must, however, account for the space occupied by sensing modules, interconnections and associated readout circuits. For example, the bare chip of a GaN multidirectional sensing device measured approximately 5.5×5.5×0.2 mm³, whereas the complete package measured 58.7×42.6×21.4 mm³. The packaged dimensions therefore provide the more relevant measure of installation space~\cite{shi2026multi,kim2026modular}. LEDs can also serve as receivers. In a finger-base displacement sensor, two LED-equipped circuit boards faced each other vertically, suspended in a transparent elastomer. Force-induced displacement changed the signal of the receiving LED. This design required no amplification circuit and had a packaged diameter of 27 mm and a height of 20 mm. Three-axis forces were estimated from the raw signals by a learning model, with mean errors of 0.05–0.07 N~\cite{elazizi2025compact}. An earlier tactile pad used a single 4 × 4 matrix of infrared LEDs under pliable foam as both emitters and detectors and scanned the matrix through multiplexed signal lines, so that a point force and its position could be calculated from two LEDs~\cite{rossiter2005novel}.

Remote readout architectures connect contact-sensitive regions to external light sources or detectors through optical fibers, allowing some rigid equipment to be located away from compliant fingers or confined spaces. Intensity readout generally requires a light source, photodetector and acquisition circuit. Examples include a plasmonic optical-fiber sensor connected to a 532 nm laser module and photodetector through multimode optical fibers, and a soft waveguide in a deep-sea gripper connected to remote optoelectronic components through plastic optical fibers (Figure~\ref{fig:3}c). Spectral or along-fiber demodulation additionally requires appropriate frequency-sweeping, interferometric and processing equipment. For example, the OFDR system for a weak-FBG skin comprised a TSL-570 tunable laser, main and auxiliary interferometric optical paths, photodetectors and balanced detectors, a digital storage oscilloscope and subsequent computation (Figure~\ref{fig:3}b-ii). Dual-comb readout additionally requires corresponding light sources and a reference branch (Figure~\ref{fig:3}b-iii), while low-intensity luminescence events are measured using a photon-counting head and counting unit (Figure~\ref{fig:3}d). Remote placement reduces hardware at the contact interface but requires space for the external instruments and fiber routing. System compactness therefore depends on the sensing head, connections and readout equipment together~\cite{luo2026self,teeple2018soft,shang2023soft,guo2023flexible,fu2026double,yao2022single}. Some systems also use light solely for signal transmission. In one such system, the output of a piezoresistive micro-electro-mechanical system (MEMS) three-axis force sensor was converted by an on-module microcontroller into pulse-width-modulated LED optical signals, which were transmitted in parallel to a processing unit through plastic optical fibers. Contact transduction in such systems remains piezoresistive, with light playing no role in transduction, and therefore falls outside the scope of optical tactile sensing discussed here~\cite{ascari2007miniaturized}.

\subsection{Point, Array and Distributed Coverage}\label{sec:3.2}

Point probes concentrate loads within relatively well-defined contact regions and are suitable for local force measurement and palpation (Figure~\ref{fig:3}e-i). Sensing area can be expanded either through discrete arrays with independent sensing elements or by arranging multiple detection channels within a continuous elastic medium. The former provides discrete spatial references through element positions; the latter estimates contact locations between elements from overlapping multichannel responses induced by a single load (Figure~\ref{fig:3}e-ii). Ruffini-type FBG skins achieve spatial encoding through strain propagation within a continuous medium (Figure~\ref{fig:3}e-iii), and modular optical platforms employ independent sensing elements~\cite{massari2022functional,ma2026low,cirillo2021tactile,kim2026modular}.

Distributed architectures allow a single optical fiber or lightguide to cover multiple sensitive regions. FBGs identify grating positions through different reflection wavelengths (Figure~\ref{fig:3}b-i), OFDR localizes sensitive regions using along-fiber reflected signals, and SLIMS uses differently dyed regions for color encoding. These approaches provide spatial references through wavelength, optical path length and spectral features, respectively. The existing RGB configuration of SLIMS can decouple multiple bending responses across three regions, indicating that recoverable deformation information is closely related to independent spectral observations~\cite{heo2006tactile,bai2020stretchable,li2026continuous}. A gradient-color-segmented optical fiber instead incorporated segments of the same color but different dopant concentrations within a single core. Under bending or compression, losses in the absorption bands of individual segments changed, while the non-absorption bands remained largely unchanged and could serve as multiwavelength references. When installed on robotic fingers and the back of a robotic hand, the fiber was used to measure joint bending angles and positions and to localize pressing~\cite{yan2025soft}.

Optical Lace embeds lightguides within a three-dimensional lattice, where structural deformation changes optical coupling between different light-guiding paths (Figure~\ref{fig:3}e-iv). Its spatial information arises from mechanical connections within the lattice and multichannel optical responses, which can be used to estimate local loads or overall shape. Unlike local surface sensing elements, these structures embed sensing functions within the load-bearing body, with network geometry and loading conditions affecting response relationships and corresponding calibration models. Point probes focus on local load measurement, surface arrays on the distribution of surface contact, and three-dimensional sensing networks on overall deformation estimation~\cite{xu2019optical}.

\subsection{Shared Readout for Multiple Tactile Points}\label{sec:3.3}

Multiple tactile points can share wiring, light sources or detection equipment through signal multiplexing, reducing the need for independent readout components. Different sensitive regions are distinguished through sequential LED activation and synchronous sampling, reflection wavelengths or along-fiber positions. The multiplexing method also affects the acquisition sequence and update frequency of signals from individual regions, thereby influencing the interval between complete tactile data acquisitions by the robot~\cite{heo2006tactile,li2026continuous,kim2026modular}.

A time-division optoelectronic platform sequentially illuminates the LEDs of individual modules and reads the corresponding signals through a shared analog channel, simplifying wiring across multiple modules (Figure~\ref{fig:3}a-ii). If the equivalent excitation, switching and sampling time per node is $\tau$ and the fixed overhead is $T_0$, the acquisition time for one cycle across $N$ nodes is approximately $T\approx N\tau +T_{0}$, and the full-array acquisition frequency is $f\approx 1/T$. In the scaling tests reported in Ref.~\cite{kim2026modular}, increasing the number of nodes from 100 to 1000 reduced the full-array sampling frequency from 380 Hz to 38 Hz, corresponding to an increase in cycle time from approximately 2.63 ms to 26.3 ms. Acquisition time thus increased approximately in proportion to the number of nodes. This relationship describes sequential acquisition alone; an actual tactile information update adds communication, demodulation and state estimation time. With pipelined processing, the sustained update frequency is limited by the slowest stage~\cite{kim2026modular}. An earlier optoelectronic skin used a similar scanning readout scheme: 36 modules containing 144 elements were read through 12 analog-to-digital conversion channels and 12 digital I/O channels, requiring 25 wires including ground. The 144 signals were transmitted to a host computer at up to approximately 380 Hz~\cite{cirillo2014artificial}. Serial buses serve the same purpose in large skins: modules of photo-reflectors under urethane foam communicated through a serial bus to reduce wiring and could be cut and pasted onto the curved surfaces of a humanoid robot~\cite{ohmura2006conformable}.

Spectral and along-fiber multiplexing arrange multiple sensitive points along a single optical fiber and share optical interrogation equipment. Conventional wavelength-division multiplexed FBGs identify individual gratings through different Bragg reflection wavelengths, with the readout system distinguishing their responses through wavelength demodulation. The reported weak-FBG skin used OFDR to spatially distinguish gratings with identical central reflection wavelengths and extract the spectral information associated with each grating. OFDR reconstructs along-fiber optical responses through frequency sweeping, interferometric signal acquisition and correlation calculations, which are then used to estimate contact states. High-frequency sampling of raw optoelectronic signals serves to reconstruct spectral or along-fiber information, while tactile state updates additionally require demodulation and estimation~\cite{heo2006tactile,fu2026double,li2026continuous}. Gratings embedded in exoskeletal robot end effectors were interrogated along a single fiber at kilohertz rates for closed-loop force control~\cite{park2009exoskeletal}. A single long chirped grating can encode position, because each axial position reflects a different wavelength. In one such grating, a local load opened a notch in the reflection spectrum whose central wavelength depended on the load position and strain, with sensitivities of 1.04 dBm/N for point contact and 1.87 dBm$\cdot$mm/N for a 15 mm-wide contact~\cite{li2022long}. Specific acquisition times and feedback delays are discussed in Section~\ref{sec:5}. A heterogeneous FBG biomimetic skin employed another shared readout approach. Four measurement gratings were connected sequentially through an optical switch. The overlap area between the reflection spectra of each measurement grating and a matched demodulation grating changed with peak displacement, allowing intensity-based readout by a photodetector. When both gratings were in the same temperature environment, temperature-induced shifts in the same direction produced little change in their spectral overlap area~\cite{liu2026heterogeneous}.

\subsection{Flexible Encapsulation and Environmental Adaptation}\label{sec:3.4}

Encapsulation protects optical paths and determines how contact loads are transmitted to sensitive structures. Encapsulation thickness, sensing-element embedding depth, material modulus and interfacial contact conditions affect local deformation and coupled responses between neighboring channels, thereby altering spatial sensing capability and calibration relationships. Temperature, liquid environments and mechanical loading history can also affect material and optical-path states. The curvature channel of an underwater soft-waveguide finger remained stable under high hydrostatic pressure, whereas its contact-force channel was affected by low temperature. This indicates that different measurement channels within the same device can exhibit different environmental sensitivities~\cite{teeple2018soft,gao2026s}. Surgical instruments also require compatibility with appropriate sterilization protocols, including steam autoclaving~\cite{puangmali2008state}.

An octopus-inspired soft arm integrated optoelectronic sensing, compliant suction cups and embedded connection structures, using local contact signals to support underwater adhesion and motion coordination. In the experiments, the mean force-response slopes of the device were similar in dry and wet environments, but hysteresis varied with environmental conditions and loading amplitude. The response slope characterizes the signal conversion relationship during loading, whereas hysteresis reflects the dependence of loading and unloading responses on loading history~\cite{deldottore2026peripheral}.

Large-area tactile skins must also account for connection reliability and response consistency. Modular structures facilitate replacement of local elements, whereas continuous encapsulation requires effective load transmission and optical response relationships to be maintained after repair~\cite{kim2026modular}. Connection maintenance, mechanical durability and recalibration are discussed further in the sections on calibration transfer and technological outlook. In a self-healing lightguide network, optical transmission was restored across realigned cut faces, and material self-healing recovered mechanical integrity. The authors reported drift-free and hysteresis-free dynamic sensing at 140\% strain. A soft quadruped robot protected by this network could detect cuts in its legs and adjust its gait accordingly~\cite{bai2022autonomous}.

\adrrfig{fig:3}{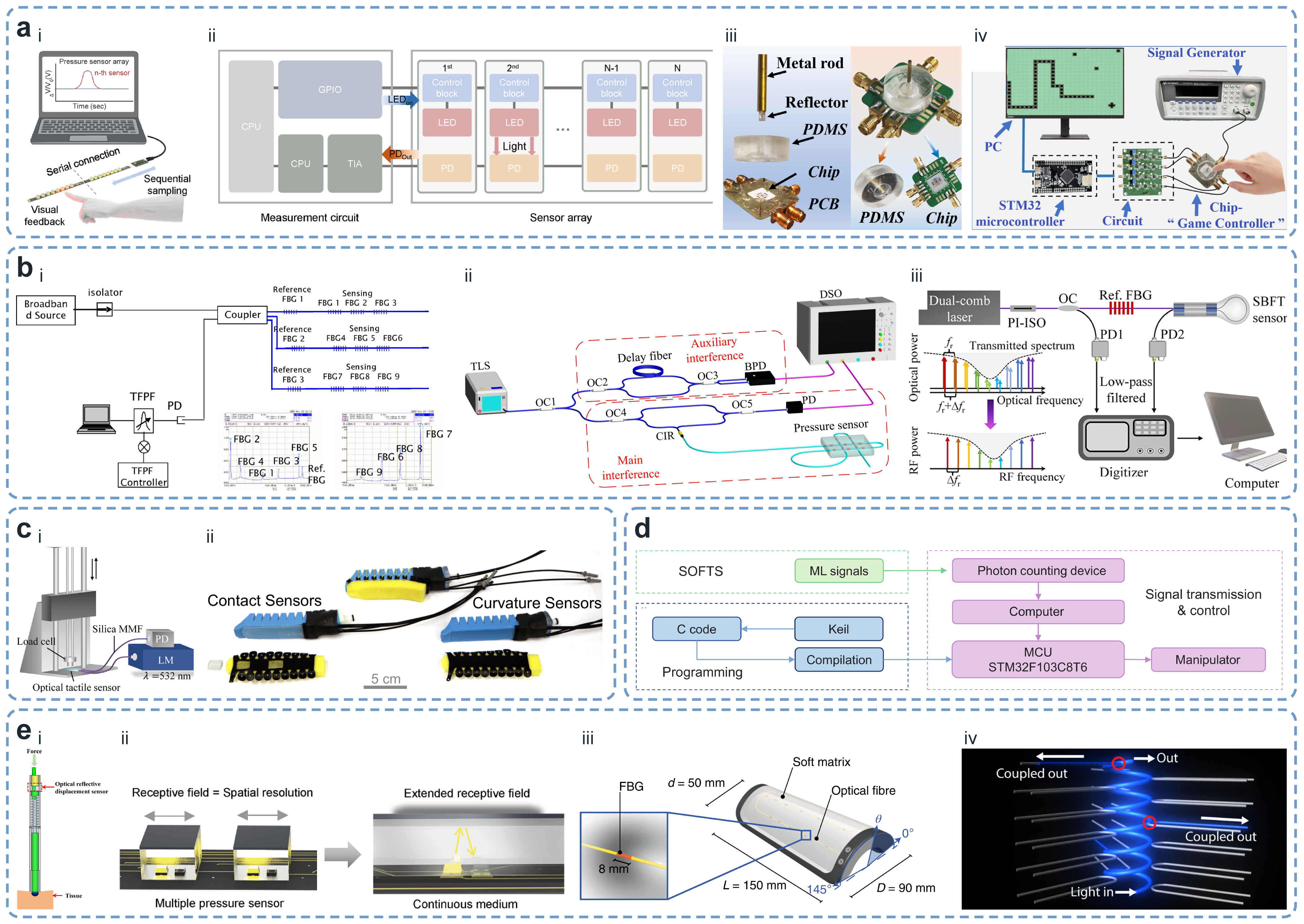}%
  {Tactile structures and readout systems in non-imaging optical tactile sensing.}%
  {%
\capline{a) Local optoelectronic readout and time-division sharing: i) modular pressure sensor array with sequential sampling; ii) measurement circuit with shared control lines and one amplifier channel. Reproduced with permission~\cite{kim2026modular}. Copyright 2026, Elsevier. iii) structure of a monolithic GaN tactile device; iv) its acquisition and control system. Reproduced with permission~\cite{shi2026multi}. Copyright 2026, Optica Publishing Group.}
\capline{b) Spectral and along-fiber readout: i) wavelength-division-multiplexed 3 × 3 fiber Bragg grating (FBG) array. Reproduced with permission~\cite{heo2006tactile}. Copyright 2005, Elsevier. ii) optical frequency-domain reflectometry (OFDR) system for identical-wavelength weak FBGs. Reproduced under the terms of the Optica Open Access Publishing Agreement~\cite{fu2026double}. Copyright 2026, Optica Publishing Group. iii) dual-comb laser interrogating a macrobent FBG sensor. Reproduced with permission~\cite{shang2023soft}. Copyright 2023, American Chemical Society.}
\capline{c) Remote intensity readout: i) gold-nanoparticle-doped PDMS fiber sensor linked by multimode fiber to a 532 nm laser and a photodetector. Reproduced with permission~\cite{guo2023flexible}. Copyright 2023, Wiley-VCH. ii) soft-waveguide sensors on deep-sea gripper fingers, connected by optical fibers to external optoelectronics (scale bar, 5 cm). Reproduced with permission~\cite{teeple2018soft}. Copyright 2018, IEEE.}
\capline{d) Photon counting of mechanoluminescence (ML) in a robotic grasping system. Reproduced with permission~\cite{luo2026self}. Copyright 2025, Wiley-VCH.}
\capline{e) Coverage, from point probes to networks: i) handheld optical stiffness probe. Reproduced under the terms of the CC BY license~\cite{ma2026low}. Copyright 2026, The Authors, published by Wiley-VCH. ii) isolated sensor modules versus a continuous optical medium. Reproduced with permission~\cite{kim2026modular}. Copyright 2026, Elsevier. iii) continuous skin with an embedded FBG fiber. Reproduced under the terms of the CC BY 4.0 license~\cite{massari2022functional}. Copyright 2022, The Authors, published by Springer Nature. iv) three-dimensional optical lace network. Reproduced with permission~\cite{xu2019optical}. Copyright 2019, The American Association for the Advancement of Science.}%
  }

\section{Tactile Information Decoupling and Reconstruction}\label{sec:4}

This section addresses the first consideration, the identifiability of contact variables. Decoupling and reconstructing optical tactile information rely on differences in the responses induced by different contact states. The amplitude of a single optical signal may be affected simultaneously by contact force, loading location, temperature or self-bending. Multichannel outputs or spectral features can provide additional observables to distinguish these factors. The mapping between contact variables and optical signals provides the basis for structural compensation, physics-based calibration and data-driven reconstruction.

\subsection{Contact Variables and Signal Mapping}\label{sec:4.1}

When contact location and geometric conditions are known, loads can be estimated from a small number of optoelectronic outputs through calibration. When contact location, loading direction and self-deformation vary simultaneously, differences in their effects on observed signals must be used to distinguish these variables. The output of intensity-based waveguides is affected by transmission loss and local deformation (Figure~\ref{fig:4}a-i), while grating peak shifts contain contributions from both contact-induced strain and temperature. Distinguishing contact regions from reference regions, arranging different force-transmission directions or exploiting spectral features with different sensitivities can increase the number of independent observations available for tactile decoding~\cite{zhao2016optoelectronically,shang2023soft,chen2023polymer}. A pneumatic artificial muscle carried two reflective diaphragms, one responding mainly to pressure and the other to length. Calibration mapped the two LED–photodiode signals to contraction force and length, with mean end-effector force errors of 0.16 N in static and 0.12 N in dynamic tests~\cite{tiziani2020optical}. Six crossed stretchable waveguides with LEDs embedded in their cores distinguished normal indentation from tangential loading through the pattern of losses across the guides and separated incipient from gross tangential slip~\cite{fliegans2026multiple}.

Different target mechanical quantities require different measurement references. Pressure measurement must account for the loaded area, stiffness estimation requires the relationship between force and displacement, and recognition of known material categories can use features of contact responses. A handheld dual-channel probe measured contact force and relative displacement separately and estimated object stiffness after correcting for the compliance of the probe itself. Soft materials were distinguished by a micro/nanofiber hardness probe through differences in their responses under controlled contact~\cite{tang2021optical}.

\subsection{Structural Compensation and Multiparameter Decoupling}\label{sec:4.2}

Tactile decoupling can begin with the design of force-transmission structures and optical-path layouts. A self-calibrating polymer optical-fiber device used a double-layer structure and a bottom light-leakage channel to obtain observations related to contact size, indentation depth and three-axis forces (Figure~\ref{fig:4}b-i,ii). Separate curvature and local contact channels in an underwater soft finger (Figure~\ref{fig:4}a-ii) allowed self-deformation information to compensate for contact responses. A nine-waveguide gripper combined bending, contact and load signals for sensing. All these designs constrain tactile estimation by increasing the number of available observations~\cite{teeple2018soft,chen2023polymer,fan2026intelligent}. In an optical-fiber tendon-driven underactuated robotic hand, three FBGs were arranged along a single optical fiber that doubled as a finger-flexion tendon. The grating in the actuation section measured tendon tension. The grating on the distal finger pad was positioned beyond the bonding point and isolated from actuation tension, so it responded only to fingertip contact. The grating on the back of the finger was mechanically isolated to provide a temperature reference. All three gratings were read out by the same interrogator to provide force and tactile feedback during teleoperation~\cite{yi2026multifingered}. A soft cilia array incorporated both an optical channel, which measured changes in transmitted light using a photoresistor, and a cracked-graphite strain channel within pneumatic cilia, using redundant proprioceptive sensing to infer its own state and external contact in real time~\cite{yu2026integrated}.

Multiband and dual-resonance structures can exploit differences in optical response sensitivities to achieve multiparameter decoupling. A single microfiber measured temperature through a short-wavelength cut-off peak and pressure through long-wavelength optical intensity (Figure~\ref{fig:4}c-i,ii). The different sensitivities of grating and WGM resonances to temperature and load were used for joint inversion in a macrobend FBG (Figure~\ref{fig:4}c-iii). Denoting the observation vector by $\mathbf{z}$ and the target quantities by $\mathbf{q}$, their local linear relationship can be approximated by Equation~\eqref{eq:3}:
\begin{equation}
\Delta \mathbf{z}=\mathbf{K}\Delta \mathbf{q}
\label{eq:3}
\end{equation}
where $\mathbf{K}$ is the calibrated sensitivity matrix. Multiparameter decoupling requires the number of observables to be no smaller than the number of variables to be estimated and the sensitivity matrix to have full column rank. For two observables and two variables, this is equivalent to a non-zero determinant. When the response columns are nearly proportional, inversion amplifies noise. Whether an additional channel provides more identifiable information depends on its relative sensitivity; estimation accuracy further depends on noise and its correlations. This local linear relationship applies within the temperature, load and structural conditions covered by joint calibration~\cite{shang2023soft,yao2022single}. An S-shaped micro/nanofiber chip jointly estimated microcontact force and temperature using two interference dips. Their force sensitivities were −102.14 and −86.43 pm/mN and their temperature sensitivities −115.76 and −102.24 pm/°C, so the two response columns were nearly proportional~\cite{yang2025bionic}. The team that proposed the single-microfiber dual-parameter approach subsequently used separate optical fibers to measure force and temperature in a nanofiber skin and described the limitations of simultaneous measurement using a single fiber. When the contact region extended beyond the 2–3 mm-long waist region, loading of the tapered region also shifted the temperature peak. When the load exceeded 10 N, severe attenuation of higher-order modes caused the temperature peak to disappear. Simultaneous measurement also required a white-light source and spectrometer, making integration into robotic systems difficult~\cite{tang2023optical}. A coaxial hydrogel optical fiber instead incorporated two separate physical channels within one fiber: pressure was measured through light-guiding loss in a U-shaped fiber, and temperature through temperature-dependent changes in the resistance of an ionic hydrogel. The device maintained low cross-sensitivity when temperature varied over −20 to 60 °C and pressure over 0 to 60 kPa at the same time~\cite{zhou2026coaxial}.

Multi-axis tactile sensing likewise relies on distinguishable responses to different loading directions. Quadrant structures characterize displacement direction through differential channel signals, asymmetric microfibers provide directional information through differences in optical loss under different loading directions, and GaN multidirectional devices combine local optoelectronic channels with force-transmission structures to acquire multidirectional contact responses (Figure~\ref{fig:4}b-iii). Differential channels and asymmetric structures encode the direction of mechanical action into optical outputs, but their mapping relationships change with preload, contact location and encapsulation state~\cite{khamis2019novel,shi2026multi,xie2024optical}. A soft three-axis optical force sensor incorporated a light-blocking structure between two LEDs and four photodiodes that moved with deformation of the housing. Normal force displaced the structure downward, blocking more light, whereas tangential force displaced it laterally, producing asymmetric changes in optical intensity across the channels. When a gated recurrent unit (GRU) network was used to decouple the three force components, the errors in the x, y and z directions were 5.41\%, 4.85\% and 1.62\% of full scale, respectively~\cite{mo2026highly}. A medical end-effector sensor arranged five FBGs in a layered elastic structure, and structural self-decoupling together with a temperature–force model gave an interaxis coupling as low as 0.01\% of full scale within $\pm$5 N~\cite{liang2025novel}.

\subsection{Physics-Based Calibration and Data-Driven Reconstruction}\label{sec:4.3}

When the sensing structure has well-defined geometry and the number of target variables is small, linear mapping, polynomial regression or mechanical models can be used to convert optical outputs into forces and displacements. Quadrant-sensing soft pillars used multivariable regression to estimate contact forces and displacements separately, while a dual-channel stiffness probe combined optoelectronic calibration with correction for probe compliance to estimate object stiffness (Figure~\ref{fig:4}d). These methods incorporate structural parameters and contact references into signal mapping, facilitating interpretation of the response relationships of local sensing elements and their applicable operating ranges~\cite{ma2026low,khamis2019novel}.

Continuously encapsulated structures and high-dimensional spectral readout can reconstruct tactile information from multiple response features, but different tasks impose different requirements on generalization. A Ruffini-type FBG skin continuously estimated force and location and held out complete indentation events for testing, evaluating prediction performance on new contact events on the same skin. A polymer optical-fiber skin sampled 49 predefined locations and then randomly split continuous recordings, so the resulting errors reflected prediction performance under sample-level splitting. Prediction on independent events or unsampled locations requires separate evaluation. An MZI tactile pad treated location, force and shape as predefined categories and tested location and force recognition using subsequent loading cycles. Continuous estimation requires examining whether the mapping remains valid as contact conditions change, whereas category recognition is limited by the predefined label set. Comparisons among these approaches must therefore distinguish test partitioning, changes in contact conditions and output tasks. Classification accuracy for known shapes cannot be equated with the ability to reconstruct unknown shapes~\cite{massari2022functional,leong2024artificial,li2024learning}.

Different input representations can organize spatial and temporal features in optical signals. OFDR cross-correlation spectra preserve spectral-shape information from local along-fiber responses (Figure~\ref{fig:4}e-i,ii), recurrence plots represent recurring structures in channel time series (Figure~\ref{fig:4}e-iii), and spectral or wavelet features of scanning vibrations reflect dynamic changes during contact. Compared with calibration based on a single amplitude, these methods exploit richer signal features for spatial contact reconstruction, tactile state recognition and surface exploration, respectively~\cite{li2026continuous,keser2021fiber,lyu2023tactile}. A three-finger FBG robotic gripper instead used strain time series from three gratings during grasping directly as network inputs. Its long short-term memory (LSTM) network included a squeeze-and-excitation module to strengthen interchannel relationships and reached a classification accuracy of 95.97\% for six classes of objects with similar sizes and shapes~\cite{lyu2022three}. A triaxial FBG needle sensor for percutaneous puncture was decoupled with a backpropagation neural network, reaching resolutions of 2.6 mN axially over 8 N and 0.25 mN radially over $\pm$1 N with a crosstalk of 4.36\%~\cite{xie2025design}.

\subsection{Applicability and Adjustment of Calibration Relationships}\label{sec:4.4}

Calibration maps optical outputs to force, location or contact events under specified loading, mounting and environmental conditions. Changes in contact geometry, encapsulation, curvature or temperature can alter both the optical baseline and the mechanical load path, affecting the accuracy of an existing mapping. Calibration transfer therefore involves identifying which response characteristics remain stable and which require correction. The following examples consider changes in contact location, device, temperature and mounting.

Existing studies have examined the applicability of calibration relationships under changes in contact location, device and environmental temperature. For a weak-FBG skin, reconstruction was tested at contact locations excluded from training, which evaluates the model's ability to estimate new locations on the same skin (Figure~\ref{fig:4}f-i). PapillArray applied a slip model trained on one device version to online detection using another version, demonstrating cross-device applicability for a specific task. To address temperature-induced changes in interference spectra, an MZI study improved recognition performance by incorporating data acquired at different temperatures into training, enabling the model to handle temperature variations covered by the training data (Figure~\ref{fig:4}f-ii)~\cite{fu2026double,li2025integrating,wang2024robust}. Other studies compensated temperature with a reference channel or a physical model: a reference grating did so in an FBG touch–slip sensor for prosthetic hands~\cite{wang2024fiber}, and a thermo-mechanical coupling model removed the temperature-induced baseline drift of an interferometric bronchoscopy sensor~\cite{wang2026fiber}.

Remounting can change both the optical baseline and the mechanical load-transmission relationship. The alignment of soft-waveguide ends affects the initial optical intensity. A near-infrared reflective device attached to the back of a human finger used skin deformation and changes in blood volume to modulate tissue scattering and absorption, with responses varying according to attachment location and user. After each attachment, the latter device collected new calibration data using reference force measurements and retrained a convolutional neural network (CNN) to restore pressure estimation (Figure~\ref{fig:4}f-iii)~\cite{teeple2018soft,ando2026flexible}.

The appropriate adjustment depends on how the response has changed. Zero-point and gain corrections are suitable for repeatable offsets or scaling changes when the normalized response shape, location dependence and interchannel coupling remain stable. Changes in the load path or nonlinear channel relationships instead call for calibration data under the new conditions and an updated mapping. Independent contact measurements can then assess the recovered force and location accuracy. Some designs reduce the need for recalibration by construction: red and blue light-absorbing layers on either side of a PDMS waveguide core gave the bending angle and direction for joint radii of 10–18 mm without additional calibration~\cite{zhong2025calibration}. Others build loading-rate dependence into the calibration itself, as in a rate-dependent support vector regression model for an intensity-modulated surgical grasper sensor~\cite{bandari2020miniaturized}.

\subsection{Active Pressing and Surface Scanning}\label{sec:4.5}

Active pressing and surface scanning can acquire contact loads and sense object surface shape, softness and texture. Under comparable pressing conditions, differences in contact responses can be used to compare object softness. During surface scanning, temporal signals generated by changes in contact state can be used to recognize shape or texture. A soft optical-waveguide hand sensed object softness and surface shape through predefined pressing and scanning actions (Figure~\ref{fig:4}g-i), while an FPI probe converted diaphragm vibrations during sliding into optical intensity time series and used their temporal features to recognize surface texture~\cite{zhao2016optoelectronically,keser2021fiber}.

Exploratory actions directly affect the formation of tactile features. Scanning speed, normal load and motion trajectory alter contact vibrations and material deformation, causing the same object to produce different optical time series under different conditions. An FPI probe exhibited different texture recognition performance under controlled conveyor-belt and manual scanning conditions (Figure~\ref{fig:4}g-ii–iv), indicating that the relationship between optical signals and texture features is influenced by exploratory actions. By controlling or recording these conditions, active contact provides a motion reference for inferring object characteristics from optical responses~\cite{keser2021fiber}.

\adrrfig{fig:4}{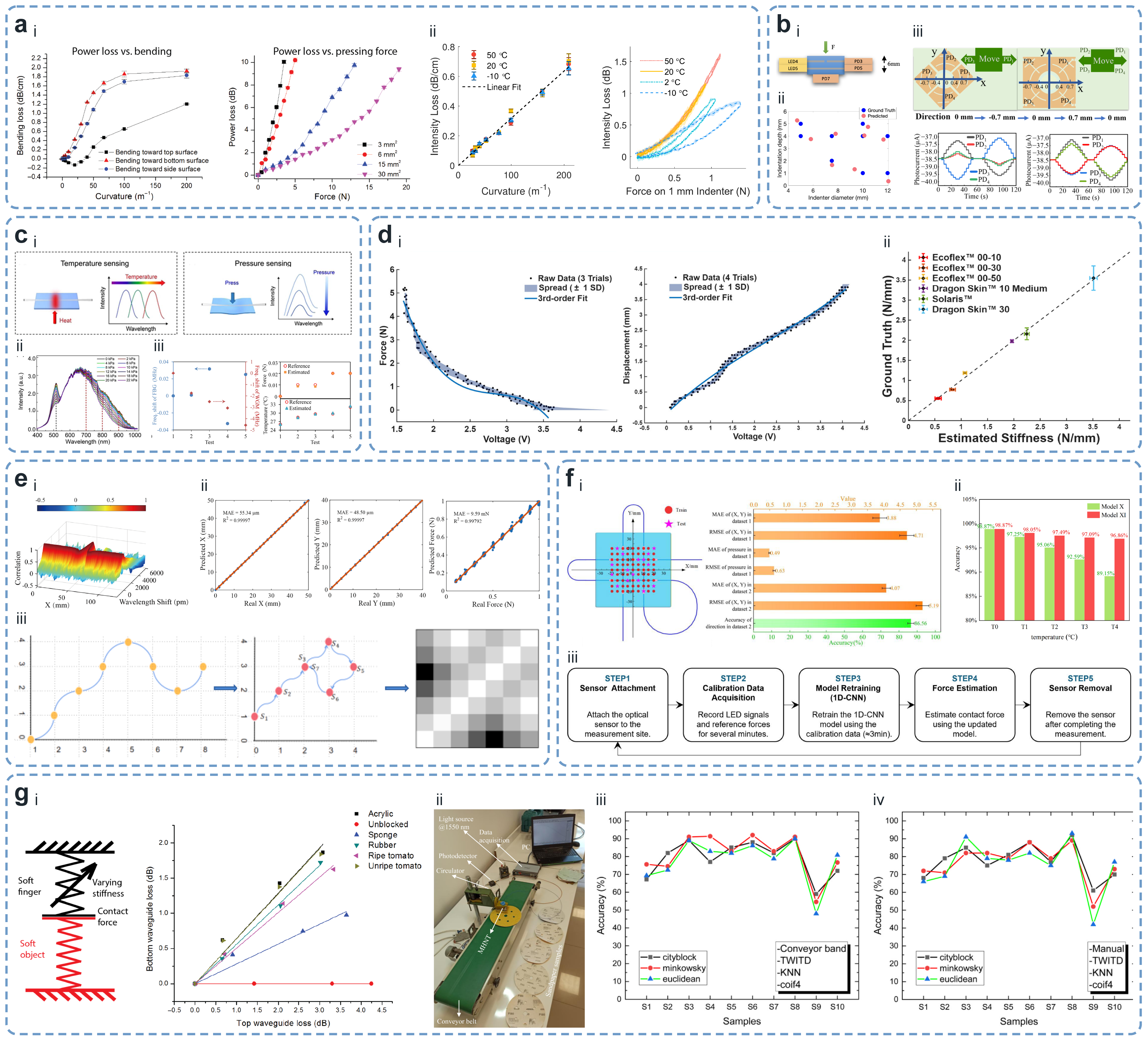}%
  {Decoupling and reconstruction of tactile information in non-imaging optical tactile sensing.}%
  {%
\capline{a) Several inputs in one optical output: i) power loss of a stretchable waveguide under bending and under pressing. Reproduced with permission~\cite{zhao2016optoelectronically}. Copyright 2016, The Authors, some rights reserved; exclusive licensee AAAS. ii) curvature and contact-force sensors of a deep-sea gripper at −10 to 50 °C. Reproduced with permission~\cite{teeple2018soft}. Copyright 2018, IEEE.}
\capline{b) Structural differential and reference channels: i) self-calibrated polymer optical-fiber sensor (side view); ii) predicted versus real indentation depth and indenter size. Reproduced with permission~\cite{chen2023polymer}. Copyright 2023, IEEE. iii) photocurrents of four receivers of a multidirectional GaN device under rod displacement. Reproduced with permission~\cite{shi2026multi}. Copyright 2026, Optica Publishing Group.}
\capline{c) Multiband and dual-resonance readout: i) single optical microfiber in PDMS; ii) transmission spectra under 0–22 kPa. Reproduced with permission~\cite{yao2022single}. Copyright 2022, Chinese Laser Press. iii) FBG and whispering-gallery-mode resonance shifts and the decoupled pressure and temperature. Reproduced with permission~\cite{shang2023soft}. Copyright 2023, American Chemical Society.}
\capline{d) Physical calibration of a handheld stiffness probe: i) force–voltage and displacement–voltage calibrations; ii) estimated versus ground-truth stiffness. Reproduced under the terms of the CC BY license~\cite{ma2026low}. Copyright 2026, The Authors, published by Wiley-VCH.}
\capline{e) Learning from spectral shape and time series: i) optical frequency-domain reflectometry (OFDR) cross-correlation spectrum of a micro/nanofiber skin; ii) location and force predicted by a ResNet-18 network. Reproduced with permission~\cite{li2026continuous}. Copyright 2026, Wiley-VCH. iii) recurrence-plot encoding of multichannel FBG signals for a VGG classifier. Reproduced with permission~\cite{lyu2023tactile}. Copyright 2023, Elsevier.}
\capline{f) Validity of a calibration under changes in position, temperature and attachment: i) errors at untrained locations on a double-layer IWFBG skin. Reproduced under the terms of the Optica Open Access Publishing Agreement~\cite{fu2026double}. Copyright 2026, Optica Publishing Group. ii) recognition accuracy of an MZI tactile sensor at different temperatures. Reproduced under the terms of the Optica Open Access Publishing Agreement~\cite{li2025integrating}. Copyright 2025, Optica Publishing Group. iii) recalibration workflow of a finger-worn near-infrared reflective sensor. Reproduced under the terms of the CC BY 4.0 license~\cite{ando2026flexible}. Copyright 2025, The Authors, published by IEEE.}
\capline{g) Active pressing and surface scanning: i) softness detection with a soft prosthetic hand. Reproduced with permission~\cite{zhao2016optoelectronically}. Copyright 2016, The Authors, some rights reserved; exclusive licensee AAAS. ii) fiber-optic Fabry–Pérot surface-roughness probe; recognition accuracy for ten sandpapers under iii) conveyor-belt and iv) manual scanning. Reproduced with permission~\cite{keser2021fiber}. Copyright 2021, Elsevier.}%
  }

\section{Performance Characteristics and Trade-Offs}\label{sec:5}

The performance of optical tactile sensors is jointly determined by transduction mechanisms, mechanical structures and signal readout. Different technological approaches have distinct characteristics in terms of load detection range, spatial localization capability and feedback speed. These characteristics may constrain one another through the allocation of measurement resources, but they may also be improved together through material and structural optimization. This section analyzes performance characteristics and associated trade-offs in relation to specific device configurations. Spatial coverage and feedback timing are examined at the system level (Sections~\ref{sec:5.2} and~\ref{sec:5.3}), and each reported value is kept with the test conditions that determine what it is evidence of (Table~\ref{tab:2}).

\adrrfig{fig:5}{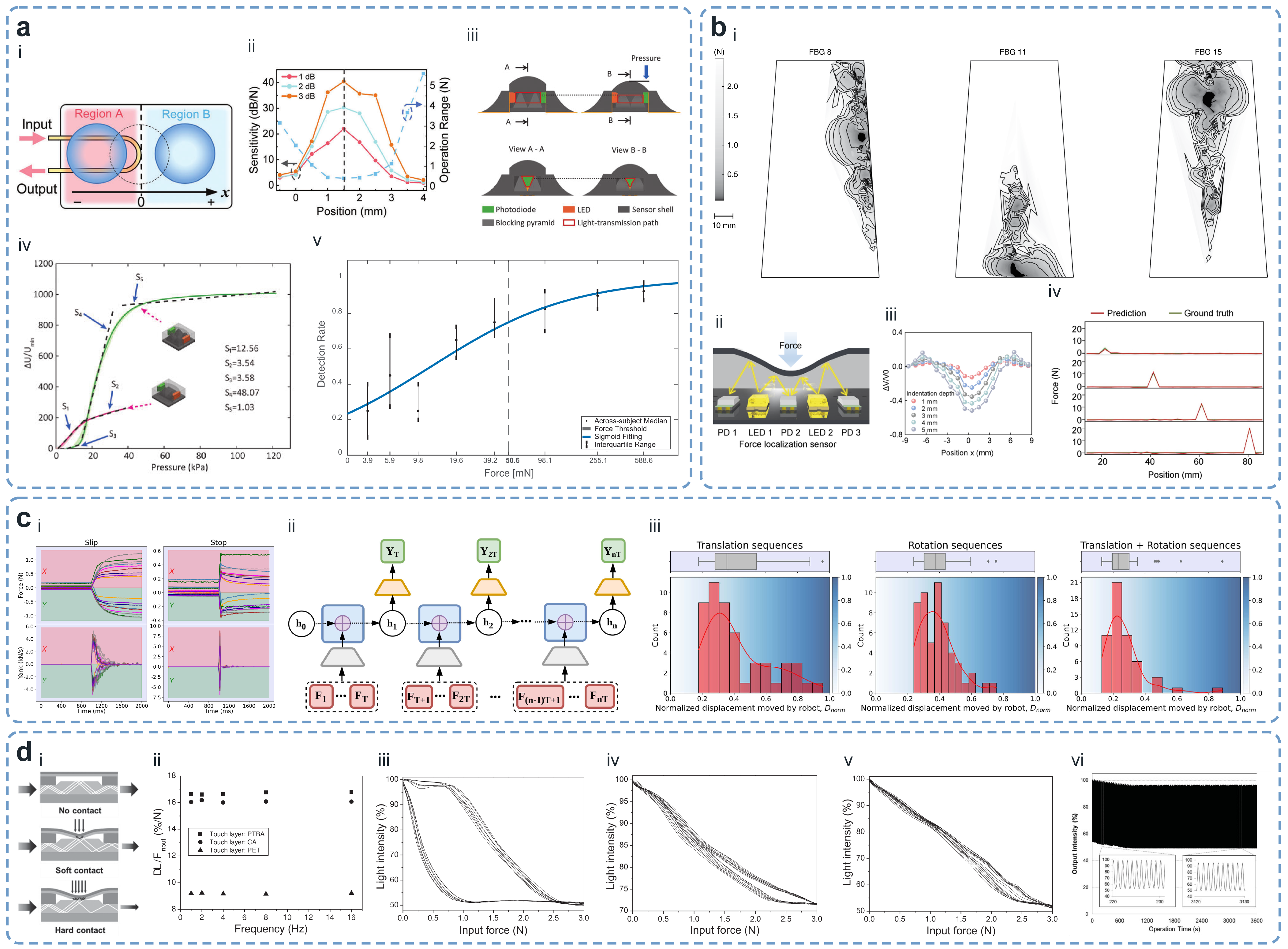}%
  {Performance characteristics and trade-offs of non-imaging optical tactile sensors.}%
  {%
\capline{a) Sensitivity, range and detection capability: i) protrusion position along a U-shaped micro/nanofiber (MNF); ii) sensitivity and operation range versus protrusion position. Reproduced under the terms of the CC BY license~\cite{xie2024optical}. Copyright 2024, The Authors, published by Wiley-VCH. iii) intentional-blocking photoelectric pressure sensor; iv) relative voltage variation versus pressure for double-row and single-row pyramids. Reproduced with permission~\cite{li2023intentional}. Copyright 2022, Mary Ann Liebert, Inc. v) touch detection rate of a large-area FBG skin versus von Frey filament force.}
\capline{b) Overlapping responses for coverage and position estimation: i) receptive fields of three FBGs in the same skin. a-v and b-i: Reproduced under the terms of the CC BY 4.0 license~\cite{massari2022functional}. Copyright 2022, The Authors, published by Springer Nature. ii) continuous optical medium over interleaved LEDs and photodiodes; iii) relative voltage responses for indentation positions and depths; iv) predicted and ground-truth force distributions. Reproduced with permission~\cite{kim2026modular}. Copyright 2026, Elsevier.}
\capline{c) From contact to action in slip detection with a PapillArray sensor: i) tangential force and its rate of change during slip and stop events; ii) windowing of the sequence data for a GRU classifier; iii) robot displacement at incipient-slip detection. Reproduced with permission~\cite{wang2024robust}. Copyright 2023, IEEE.}
\capline{d) Touch-layer materials under the same loading, and cycling durability: i) working principle of a polymer-waveguide force sensor; ii) sensitivity at 1–16 Hz for PTBA, CA and PET touch layers; iii–v) hysteresis with PTBA (iii), PET (iv) and CA (v) touch layers; vi) cycling durability with the CA touch layer. Reproduced with permission~\cite{yun2014polymer}. Copyright 2014, Wiley-VCH.}%
  }

\subsection{Sensitivity, Measurement Range and Detection Capability}\label{sec:5.1}

Sensitivity characterizes the rate of change in sensor output with respect to a target stimulus. Optical loss per unit force (e.g. dB/N), wavelength shift per unit force (e.g. pm/N) and relative photocurrent change per unit pressure (e.g. kPa⁻¹) represent different measurement forms. The relative photocurrent change uses the unloaded current $I_0$ as its baseline and is expressed as $(I-I_0)/I_0$. Because the output quantities and units differ, sensitivities cannot be ranked by their numerical values. Mechanical and optical structures jointly influence the conversion from load to output. The bending region and loading position of a U-shaped microfiber determine its optical-loss response, the contact layer and optoelectronic configuration of a planar skin affect its pressure response, and FBG encapsulation generates fiber strain through load transmission. Sensitivity is therefore a local response characteristic of the complete sensing system under specific structural and loading conditions and within a defined measurement interval~\cite{ramuz2012transparent,xie2024optical,leong2024artificial}.

Enhancing local optical responses facilitates light-load detection but may affect the effective measurement range. At x=1.5 mm, a U-shaped microfiber achieved a normal-force response slope of approximately 50.7 dB/N within a tested response depth of up to approximately 20 dB, corresponding to an operating range of approximately 0.4 N. Figure~\ref{fig:5}a-i defines the protrusion position x, and Figure~\ref{fig:5}a-ii shows the sensitivity and operating range at different response depths. Both the response slope and measurement range varied with loading position. Concentrating local deformation within an optically sensitive region can enhance the output response, but the structural design must still match the target load and contact location. Comparisons of contact-layer materials in planar waveguides also showed that material and interface design can improve dynamic recovery characteristics~\cite{yun2014polymer,xie2024optical}. The sensitivity of an occlusion-based optoelectronic pressure sensor likewise varied across measurement intervals. A single-row pyramidal structure exhibited a sensitivity of approximately 12.56 kPa⁻¹ below 15 kPa. For a double-row pyramidal structure, sensitivity was only approximately 3.58 kPa⁻¹ below 13 kPa because the photodiode was close to saturation, increased to approximately 48.07 kPa⁻¹ over 13–32 kPa and decreased to approximately 1.03 kPa⁻¹ over 32–120 kPa (Figure~\ref{fig:5}a-iii,iv)~\cite{li2023intentional}.

Detection capability depends on readout noise and evaluation criteria, not only on the response slope that converts small mechanical changes into readable signals. The best force resolution of a helical FBG array was 0.93 mN, calculated from a wavelength readout resolution of 1 pm and the sensitivity of the corresponding sensing element. A Ruffini-type skin used a wavelength change exceeding 2 pm in any FBG as its touch detection criterion, with a force threshold of approximately 50.6 mN corresponding to a fitted detection probability of 75\% (Figure~\ref{fig:5}a-v). The 0.8 mN reported for an FBG/WGM device was instead the root-mean-square error of force estimation under combined loading conditions. These three quantities characterize force resolution calculated from instrument resolution and sensor response, contact-event detection capability and multiparameter joint inversion accuracy, respectively~\cite{massari2022functional,shang2023soft,li2018high}. Optical interference can also limit detection capability. Interference-like oscillations were superimposed on the response curves of a VCSEL shear sensor, which the authors attributed to laser reflections at the photodiode surface. The lateral displacement accuracy was consequently limited to approximately 3 μm~\cite{missinne2012two}.

\subsection{Spatial Coverage and Location Estimation}\label{sec:5.2}

Spatial information can be provided by the positions of independent sensing elements or reconstructed from the combined responses of multiple channels within a continuous medium. Independent arrays establish a direct correspondence between output signals and contact regions. Overlapping receptive fields exploit continuous responses induced by loading between sensing elements to support finer single-point contact localization (Figure~\ref{fig:5}b). Under multipoint contact, load transmission through flexible structures and the superposition of optical signals may affect the ability to distinguish different contact locations. Sensing layouts and calibration methods must therefore be selected according to the anticipated single-point or multipoint contact conditions~\cite{massari2022functional,levi2013soft,kim2026modular}.

A distributed microfiber skin used along-fiber spectral shapes and overlapping responses for single-point contact localization. The original study used an augmented dataset containing 71.43\% synthetic samples and divided it into training, validation and test sets in proportions of 70\%, 15\% and 15\%. Under this protocol, the mean absolute error was 55.3 μm in the X direction and 48.5 μm in the Y direction. Its correlation window corresponded to a sensing spatial resolution of approximately 1.48 mm, and the learning model estimated contact locations from changes in multiple local responses. The window scale and the location prediction errors are different quantities and correspond to different measurement stages. These results characterize single-point prediction under the reported augmented-data split~\cite{li2026continuous}.

Tactile coverage can be expanded either by increasing the number of local sensing elements or by extracting spatial information from continuous optical structures. Modular platforms reduce wiring complexity across multiple sensing elements through shared connections, whereas continuous grating or OFDR structures distinguish sensitive locations within the same optical path through spectral or along-fiber demodulation. The relationships among coverage area, the number of independent observation channels and recoverable tactile variables depend on the specific encoding method. Spatial sensing capability therefore cannot be evaluated solely from sensing-element spacing or the number of output nodes~\cite{heo2006tactile,li2026continuous,kim2026modular}. A reticular core of intersecting waveguides in a single layer reduced waveguide thickness compared with split-level cores and gave more consistent sensitivities along the two shear directions, with an average sensitivity of 0.36 dB/N~\cite{liu2023two}. Along one fiber, a chirped grating encodes the contact position in the wavelength of a spectral notch~\cite{li2022long}.

\subsection{Dynamic Response and Feedback Time}\label{sec:5.3}

Dynamic tactile feedback requires contact changes to be converted promptly into robotic actions. Mechanical contact layers affect load transmission and deformation recovery, optical readout equipment determines signal sampling and acquisition speed, and decoding and actuation stages influence when feedback takes effect. The delay measured for a thin-film waveguide under 16 Hz sinusoidal loading reflects the dynamic response of the mechanical structure and device. The minimum spectral sampling time of approximately 1.1 ms for dual-comb readout characterizes the acquisition speed of the optical demodulation stage. These measurements correspond to different stages of the feedback chain~\cite{shang2023soft,yun2014polymer}. A coaxial hydrogel optical-fiber skin exhibited a measured response time of less than 10 ms. The authors noted that this value had reached the resolution limit of the spectrometer used and that the intrinsic device response was expected to be faster~\cite{zhou2026coaxial}. The viscosity of soft layers affects optoelectronic response. An optoelectronic skin covered with silicone exhibited a step response time of approximately 1 ms, while an occlusion-based optoelectronic pressure sensor had a response time of 33 ms and a recovery time of 96 ms under a load of 30 kPa~\cite{li2023intentional,cirillo2014artificial}.

When the stages operate approximately in series, the time required for a single feedback cycle, from a change in contact state to the onset of actuator response, can be expressed by Equation~\eqref{eq:4}:
\begin{equation}
t_{\mathrm{loop}}\approx t_{\mathrm{trans}}+t_{\mathrm{scan}}+t_{\mathrm{comm}}+t_{\mathrm{decode}}+t_{\mathrm{control}}+t_{\mathrm{act}}
\label{eq:4}
\end{equation}
The terms represent the time required for mechanical load transmission and transduction response, acquisition waiting and signal acquisition, communication, tactile decoding, control computation and the onset of actuator response, respectively. In practical systems, some stages may overlap or operate in parallel, so the total feedback delay also depends on the specific execution sequence. This decomposition helps identify the main sources of delay and the corresponding stages for optimization.

Signal multiplexing changes how acquisition time is allocated. Sequential readout in modular platforms causes the full-array sampling frequency to decrease approximately inversely with the number of nodes (Section~\ref{sec:3.3}). The raw interferometric signal sampling rate of an OFDR microfiber skin was 31.25 MS/s, but a single complete acquisition required approximately 1.05 s, followed by spectral processing and contact information reconstruction. Raw signal sampling rate, complete acquisition time and tactile feedback delay must therefore be evaluated separately. An underwater soft finger used a readout frequency of 10 Hz for low-frequency operations below 1 Hz, illustrating how acquisition frequency can be matched to the dynamic requirements of the target task~\cite{teeple2018soft,li2026continuous,kim2026modular}.

Event detection is affected by temporal windows and control cycles. PapillArray acquired force data at 1000 Hz and grouped every 40 samples into non-overlapping windows, updating the sequence at 25 Hz (Figure~\ref{fig:5}c-i,ii). The classifier processed data at approximately 40 Hz, and the gripper servo refresh rate was approximately 62 Hz. High-frequency sampling provides temporal information about slip, while window updates, classification processing and servo control jointly influence when corrective actions are triggered~\cite{wang2024robust}.

\begin{table*}[!tbp]
\caption{Representative performance metrics and experimental conditions}\label{tab:2}
\begin{tabular*}{\textwidth}{@{}>{\raggedright\arraybackslash}p{0.160\dimexpr\textwidth-8\tabcolsep\relax}>{\raggedright\arraybackslash}p{0.136\dimexpr\textwidth-8\tabcolsep\relax}>{\raggedright\arraybackslash}p{0.268\dimexpr\textwidth-8\tabcolsep\relax}>{\raggedright\arraybackslash}p{0.260\dimexpr\textwidth-8\tabcolsep\relax}>{\raggedright\arraybackslash}p{0.176\dimexpr\textwidth-8\tabcolsep\relax}@{}}
\toprule
\textbf{Study} & \textbf{Metric} & \textbf{Value} & \textbf{Test conditions} & \textbf{Metric description} \\
\midrule
Transparent planar pressure-sensitive skin~\cite{ramuz2012transparent} & Normalized pressure-response slope & Maximum approximately 0.2 kPa⁻¹ & Inorganic source/photodiode readout; below 1 kPa; 70 mm² loading area & Local slope in the low-load range \\
Thin-film waveguide array~\cite{yun2014polymer} & Dynamic delay and hysteresis & CA 2.4 ms/6.7\%; PET 2.5 ms/10.7\%; PTBA 10 ms/45.1\% & Identical 16 Hz, 0–3 N sinusoidal loading & Material comparison under identical conditions; delay measured in sinusoidal tests \\
Handheld stiffness probe~\cite{ma2026low} & Force and displacement RMSE & 0.08 N/0.2 mm & Tests on six materials & Continuous-variable estimation errors \\
Polymer FBG skin~\cite{leong2024artificial} & Force MAE and single-axis location MAE & 0.07 N; X 0.24 mm/Y 0.21 mm & Up to 3 N; random sample splitting & Prediction errors under the corresponding data split \\
Ruffini-type FBG skin~\cite{massari2022functional} & Median errors on independent test data & Force 35 mN; location 3.2 mm & Complete indentation events held out & Median values, distinct from MAE \\
Helical FBG array~\cite{li2018high} & Best force resolution reported by the authors & 0.93 mN & Best-performing element; 1 pm wavelength resolution & Derived from instrument resolution and element sensitivity \\
U-shaped micro/nanofiber~\cite{xie2024optical} & Local normal-force response slope & 50.7 dB/N & x=1.5 mm; approximately 0.4 N range & Position-dependent response \\
Single-microfiber dual-parameter sensor~\cite{yao2022single} & Temperature and pressure response slopes & 1.205 nm/°C; 4.5\%/kPa & Temperature 25–95 °C; pressure at 900 nm, 6–20 kPa & Separate calibration conditions for the two wavelength bands \\
FBG/WGM dual resonance~\cite{shang2023soft} & Decoupling RMSE & 0.8 mN; 0.2 °C & Five sets of combined force and temperature conditions & Joint inversion errors \\
Distributed microfiber skin~\cite{li2026continuous} & MAE after data augmentation & X 55.3 μm/Y 48.5 μm; force 9.6 mN & 50×40 mm²; augmented dataset with 71.43\% synthetic samples; 70\%/15\%/15\% split after augmentation & Single-point prediction results \\
Modular continuous coverage~\cite{kim2026modular} & Mean force RMSE across output nodes & 16.0±0.955 mN & Leave-one-out across five cycles; mean across 30 location nodes; input loads 5–20 N & Node-averaged error, distinct from total force error for a single contact \\
\bottomrule
\end{tabular*}
\begin{tablenotes}
Note: Reported values are presented with their loading, readout and data conditions. Metric definitions are retained to support comparisons appropriate to each sensing task. The values are device characterization or offline estimation results; closed-loop use is summarized in Table~\ref{tab:3}.
\end{tablenotes}
\end{table*}

\subsection{Hysteresis, Drift and Repeatability}\label{sec:5.4}

The viscoelasticity and interfacial interactions of soft materials cause differences between loading and unloading responses, and load holding and recovery after unloading are time-dependent~\cite{yun2014polymer,wang2026relaxation}. Under identical sinusoidal loading conditions of 16 Hz and 0–3 N, thin-film waveguides with contact layers made of cellulose acetate (CA), polyethylene terephthalate (PET) and poly(tert-butyl acrylate) (PTBA) exhibited different response delays and hysteresis (Figure~\ref{fig:5}d-i–v), as shown in Table~\ref{tab:2}. In these experiments, the refractive index of the functional coating on the contact layer was fixed at 1.628, and mean hysteresis was calculated as the ratio of hysteresis error to full-scale output (FSO). Among the materials, CA substantially improved dynamic recovery while maintaining sensitivity comparable to that of PTBA. These results show that contact-layer materials and interface design affect both response amplitude and dynamic recovery~\cite{yun2014polymer}.

Cyclic loading history and assembly differences affect sensing stability in different ways. Transparent skins and dynamic waveguide arrays (Figure~\ref{fig:5}d-vi) exhibited changes in response during initial loading cycles before gradually stabilizing. Differences in the alignment of soft-waveguide ends caused initial optical intensity offsets between devices. The former reflects loading-history effects within the same structure; the latter is primarily associated with optical connections and assembly conditions. For replaceable modules and underwater contact units, long-term use also requires attention to changes in connection reliability, load transmission and recovery after unloading, with the scope of recalibration determined accordingly~\cite{teeple2018soft,deldottore2026peripheral,ramuz2012transparent,yun2014polymer,kim2026modular}. Existing optoelectronic devices have reported stability using different metrics. An occlusion-based optoelectronic pressure sensor exhibited approximately 0.4\% drift after 12,130 loading cycles over 0–80 kPa (approximately 10 h), with negligible hysteresis during loading and unloading over 0–120 kPa. For a 6×6 optoelectronic skin, the repeatability error was 6.77\% and the hysteresis error was 10.27\%, based on the worst-performing element~\cite{li2023intentional,cirillo2014artificial}.

Table~\ref{tab:2} summarizes the measurement results of representative devices and their corresponding test conditions. Mean absolute error (MAE) and root mean square error (RMSE) retain the definitions used in the original studies. Local response slopes, event detection thresholds and estimation errors are used to evaluate their respective sensing capabilities.

\subsection{Selecting Technical Routes from Task Requirements}\label{sec:5.5}

The trade-offs above indicate that route selection should begin with the information required by the robotic task rather than with a single performance metric. Four aspects of the task constrain the design. The first is the set of contact quantities that must be distinguished, such as normal and tangential force, contact location, slip events or temperature. The second is the timescale on which they are needed, from continuous tracking during holding to transient events that require short feedback delays. The third is the area to be covered, from a fingertip to a robot link or an entire arm. The fourth is the space, environment and cost available for the sensor and its readout. The first aspect is a requirement on identifiability, and the second and third are requirements on system-level timing and coverage. These requirements determine the mechanical structure that must convert the target quantities into distinguishable optical changes, the optical encoding that keeps them separable and the readout that delivers them at the required rate. Slip correction in grippers, for example, has used local intensity channels or luminescence events that mark the onset of slip, whereas contact on robot surfaces has been detected with multiplexed gratings or with modular strips that share wiring~\cite{wang2024robust,pan2023knot,luo2026self,massari2022functional,kim2026modular}. Table~\ref{tab:routes} summarizes the main advantages, key limitations and system costs of the technical routes discussed in Sections~\ref{sec:2}--\ref{sec:5}, together with the robotic applications in which they have been demonstrated (Section~\ref{sec:6}).

\begin{table*}[!tbp]
\caption{Technical routes of non-imaging optical tactile sensing: main advantages, key limitations and system costs, and main robotic applications}\label{tab:routes}
\begin{tabular*}{\textwidth}{@{}>{\raggedright\arraybackslash}p{0.28\dimexpr\textwidth-6\tabcolsep\relax}>{\raggedright\arraybackslash}p{0.23\dimexpr\textwidth-6\tabcolsep\relax}>{\raggedright\arraybackslash}p{0.26\dimexpr\textwidth-6\tabcolsep\relax}>{\raggedright\arraybackslash}p{0.23\dimexpr\textwidth-6\tabcolsep\relax}@{}}
\toprule
\textbf{Technical route} & \textbf{Main advantages} & \textbf{Key limitations and system costs} & \textbf{Main robotic applications} \\
\midrule
Reflection, occlusion and spot displacement (LED–photodiode or quadrant units)~\cite{khamis2019novel,shi2026multi,li2023intentional,kim2026modular} & Compact local units; directional output; simple intensity readout & Alignment-sensitive; electronics at the contact site; slower scans with many nodes & Slip correction; contact stop on robot links; proximity and tactile servoing \\
Waveguide loss and coupling (soft waveguides, lightguides, fiber knots)~\cite{zhao2016optoelectronically,bai2020stretchable,pan2023knot,fan2026intelligent,teeple2018soft} & Soft and stretchable; simple readout; color-coded location & Bending and pressing confounded; hysteresis; device-to-device offsets; temperature effects & Softness exploration; slip feedback; soft and underwater grasping \\
Localized fields (micro/nanofibers, plasmonic fibers)~\cite{xie2024optical,tang2021optical,guo2023flexible,zhang2020ultrasensitive} & High sensitivity and speed (e.g., 7~mPa, 10~μs); direction from asymmetric structures & Fragile tapers; position- and mounting-dependent response; sensitive to surroundings and bending & Hardness palpation; multi-axis and shear sensing \\
Fiber Bragg gratings with wavelength-division multiplexing~\cite{massari2019tactile,massari2022functional,park2009exoskeletal,webb2015fibre} & Many points per fiber; remote electronics; kHz interrogation & Interrogator required; limited spectral bands; temperature cross-sensitivity & Grasp force control; contact stop on robot surfaces; instrument-tip force \\
Distributed OFDR readout (weak gratings, Rayleigh backscatter)~\cite{fu2026double,li2026continuous} & Continuous coverage along one fiber & Tunable laser and interferometers; slow complete acquisition (about 1~s); transverse location relies on packaging & Contact localization on sensing skins \\
Interference and resonance (MZI, FPI, WGM, fiber rings)~\cite{li2024ultrasensitive,li2024learning,wang2026fiber,shang2023soft} & Sub-mN resolution; force–temperature separation & Temperature-sensitive spectra; spectral demodulation equipment & Instrument-tip force; texture and Braille recognition \\
Mechanoluminescence (photon counting)~\cite{luo2026self,pan2024quantifying} & No excitation light; marks contact onset and slip & Static load not read directly; photon counting needed & Event-triggered grasp actions \\
\bottomrule
\end{tabular*}
\begin{tablenotes}
Note: Entries summarize the studies discussed in Sections~\ref{sec:2}--\ref{sec:5}; numerical values are examples from single studies. Applications are detailed in Section~\ref{sec:6} and Table~\ref{tab:3}.
\end{tablenotes}
\end{table*}

\section{Robotic Applications}\label{sec:6}

Reported results on non-imaging optical tactile sensors provide evidence at three levels. Device characterization gives responses, ranges and errors measured on a test bench. Offline recognition or estimation applies a calibration or a learned model to recorded signals, and what it shows depends on how the test data were separated from the training data (Section~\ref{sec:4.3}). Closed-loop use lets the tactile signal change a robot action during the task and is judged by the task outcome. The values in Table~\ref{tab:2} belong to the first two levels. This section concerns the third, and the note to Table~\ref{tab:3} marks the systems in which the signal did not feed back into the action.

\subsection{From Contact Signals to Robotic Actions}\label{sec:6.1}

Different robotic tasks require different types of tactile information. During grasping, continuous contact feedback can be used to regulate finger closure, while initial slip detection helps correct the grasp before substantial object displacement occurs. Surface exploration uses predefined pressing or scanning actions to acquire information about object shape and material properties. Underwater soft arms use multipoint contact information to coordinate local adhesion and whole-arm motion. Depending on the task, the emphasis is on continuous contact quantities, dynamic contact events or multichannel contact distributions~\cite{zhao2016optoelectronically,deldottore2026peripheral,wang2024robust,massari2019tactile}.

The role of optical tactile sensing in robotic tasks depends on how the acquired information contributes to action decisions. Active exploration uses robot motion as a measurement condition, obtaining comparable responses through predefined contact actions for object selection or surface feature analysis. Online feedback instead adjusts finger closure, increases grip force, stops motion or controls adhesion in real time according to contact states during manipulation. The former emphasizes the repeatability of contact conditions; the latter additionally requires coordination among signal readout, state assessment and action execution. The following discussion examines specific applications of non-imaging optical tactile signals in robotic tasks involving slip correction, fragile-object grasping and underwater distributed manipulation~\cite{zhao2016optoelectronically,luo2026self,deldottore2026peripheral,wang2024robust,massari2019tactile}.

\subsection{Grasping, Slip Prevention and Dynamic Adjustment}\label{sec:6.2}

Slip prevention requires detecting the onset of grasp instability and adjusting actions promptly. PapillArray acquired tangential force time series from nine soft pillars through optoelectronic readout and identified incipient slip from asynchronous responses generated when some pillars began sliding relative to the object while others remained stationary. Its temporal model, based on a GRU, incorporated control sequences in which motion stopped abruptly before slip, distinguishing incipient slip from similar force changes caused by motion cessation. Upon detecting incipient slip, the robot stopped arm motion and increased grip force to a predefined secure-grasp level, achieving slip-triggered grasp correction (Figure~\ref{fig:5}c-iii)~\cite{wang2024robust}.

The method used data from the v1 device to train the model and deployed it on a v2 device with different dimensions and soft-pillar hardness. Six objects were selected for robotic grasping experiments. Insufficient grip force was used to induce translational slip, and grasps offset from the center of mass to induce rotational slip. Each scenario was repeated three times, giving a total of 36 trials. After the objects were lifted, the mean displacement relative to the gripper was 2.5 mm and the mean rotation angle was 1.9°~\cite{wang2024robust}.

Grasping fragile objects instead emphasizes limiting further finger closure after contact is established. An FBG robotic hand encapsulated grating-containing optical fibers within a flexible silicone layer on the surfaces of rigid fingers and used the mean shift in reflection peak wavelength across the gratings as the feedback variable (Figure~\ref{fig:6}a-i). Static control employed lower and upper thresholds. As the fingers closed, contact-induced strain increased. Finger closure stopped when the mean peak shift reached the lower threshold, and exceeding the upper threshold caused the fingers to open. In 20 grasping trials with commercial nachos at a closure-stop threshold of 0.16 nm, only one instance of breakage occurred. For objects with changing dimensions, the system further employed proportional–integral–derivative (PID) control to adjust the drive voltage and maintain the setpoint for the mean peak shift. In experiments where a linear stage changed the length of a mechanical telescopic component, the fingers adjusted their positions as the object dimensions changed, with transient peak-shift errors when the motion direction reversed~\cite{massari2019tactile}.

Other grasping systems adjust their grip using continuous contact signals or slip events. A nine-waveguide soft gripper regulated its air pump using voltage feedback from inner channels and increased the control setpoint when slip was detected. Bending signals from outer channels could compensate for coupled responses in the inner and lateral channels. After identifying object properties, the system could also apply predefined gripping parameters (Figure~\ref{fig:6}a-ii). An optical-fiber knot detected slip through friction-induced changes in optical intensity. In a demonstration in which screws were gradually added to a cup, the slip signal triggered further gripper closure to strengthen the grasp. With feedback disabled, the cup slipped from the gripper (Figure~\ref{fig:6}a-iii)~\cite{pan2023knot,fan2026intelligent}. In an experiment in which mung beans were gradually added to a container held by a granular-jamming soft gripper, irregular fluctuations in waveguide outputs during slip were used to adjust the grip force promptly. Without this feedback, the container eventually slipped from the gripper~\cite{mo2025empowering}. Plastic optical fibers embedded in a pneumatic soft gripper served both as its strain-limiting layer and as contact-based light-loss sensors read by photodiodes; the sensor drifted less than 0.1\% over 9000 cycles and supported slip detection and grip-force feedback, including for magnetic objects~\cite{mun2024multi}.

A mechanoluminescent gripper used light-emission peaks generated by mechanical stimulation to trigger action switching. When a contact peak exceeded the threshold, the gripper stopped closing and lifted the object. Upon detecting a slip peak, it increased the grip force until the luminescence signal disappeared. This event-triggered strategy supports contact and slip responses, while continuous force regulation during holding requires a sustained load signal~\cite{luo2026self}.

\subsection{Palpation, Instrument-Tip Force Sensing and Active Exploration}\label{sec:6.3}

Palpation uses controlled indentation to identify differences in local mechanical responses. Soft materials or tissue phantoms were distinguished by a micro/nanofiber hardness probe and a helical FBG array (Figure~\ref{fig:6}b-i) through differences in their responses under loading. A handheld dual-channel probe instead measured contact force and displacement separately and estimated object stiffness after correcting for deformation of the probe itself (Figure~\ref{fig:6}b-ii)~\cite{ma2026low,tang2021optical,li2018high}.

Contact-force sensing at the tips of surgical instruments is closely related to palpation. Minimally invasive instruments attenuate the surgeon's direct tactile feedback, while fiber-optic force sensing can facilitate magnetic resonance imaging (MRI)-compatible instrument design~\cite{puangmali2008state,polygerinos2010mri}. A discretely loaded beam read by optical fibers measured distributed tactile information with a single moving part under both static and dynamic loading~\cite{ahmadi2012discretely}. FBG-based three-axis sensors at instrument tips reached resolutions of 1.2, 1.1 and 2.5 mN for laparoscopic palpation~\cite{hao2025high} and 4.95 mN within $\pm$5 N in palpation and acupuncture tests~\cite{liang2025novel}, and an FBG needle sensor tracked puncture forces on ex vivo tissue with an RMSE of the resultant force not exceeding 12.3 mN~\cite{xie2025design}. In the test of the laparoscopic sensor on ex vivo porcine liver, the resultant force reached 1.66 N as the instrument tip slid across a simulated tumor and stayed between 0.40 and 0.63 N over normal tissue (Figure~\ref{fig:6}b-iii)~\cite{hao2025high}. An interferometric sensor for robotic bronchoscopy resolved 0.57–4.92 mN within $\pm$3 N~\cite{wang2026fiber}. These sensors measure the resultant force at the instrument tip rather than a contact distribution.

Active exploration acquires information about object surfaces through predefined movements. A soft optical-waveguide hand scanned and pressed objects, using optical-loss responses to sense shape and softness, and demonstrated the selection of a softer object from several tomatoes (Figure~\ref{fig:6}b-iv). In this process, tactile information was used for object measurement and selection, while scanning and pressing actions remained under open-loop control~\cite{zhao2016optoelectronically}. A reflective texture probe converted mirror movement caused by surface protrusions into optical time series, and an FPI probe did the same with diaphragm vibrations generated during sliding contact. In the FPI study, controlled conveyor-belt and manual scanning produced different texture recognition performance, indicating that motion trajectory, scanning speed and normal load all affect tactile observations. By repeatedly performing predefined exploratory actions, robots can obtain comparable contact responses at different locations or orientations and use them to characterize object surface features~\cite{keser2021fiber,mazid2006robotic}. A review of tactile sensing for object properties identified shape, surface material and object pose as the main target properties and discussed integrating touch with other sensing sources~\cite{luo2017robotic}.

\adrrfig{fig:6}{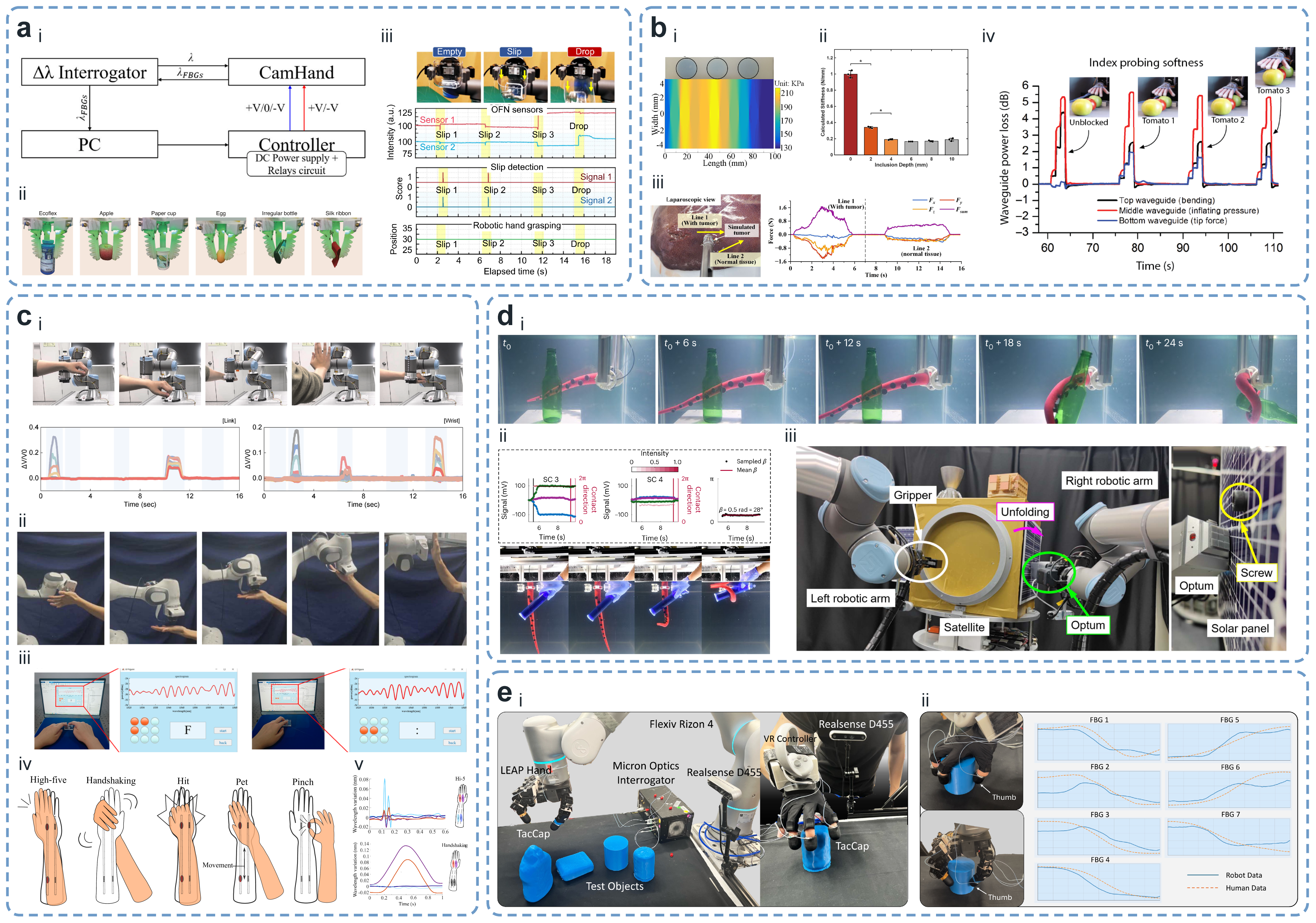}%
  {Robotic applications of non-imaging optical tactile sensors.}%
  {%
\capline{a) Grasping, slip control and dynamic adjustment: i) FBG-sensorized robotic hand with wavelength-based closing control. Reproduced under the terms of the CC BY license~\cite{massari2019tactile}. Copyright 2019, The Authors, published by Frontiers. ii) closed-loop adaptive grasping by a soft gripper with nine waveguide sensors. Reproduced under the terms of the CC BY 4.0 license~\cite{fan2026intelligent}. Copyright 2026, The Authors, published by Springer Nature. iii) grasping with the slip feedback of knot-shaped optical-fiber (OFN) sensors switched off. Reproduced under the terms of the CC BY 4.0 license~\cite{pan2023knot}. Copyright 2023, The Authors, published by the Institute of Optics and Electronics, Chinese Academy of Sciences.}
\capline{b) Palpation, instrument-tip force sensing and active exploration: i) stiffness map of a silicone phantom from a spiral-FBG palpation array. Reproduced with permission~\cite{li2018high}. Copyright 2018, IEEE. ii) stiffness measured by a handheld probe versus nodule depth. Reproduced under the terms of the CC BY license~\cite{ma2026low}. Copyright 2026, The Authors, published by Wiley-VCH. iii) laparoscopic view and three-axis forces of an FBG sensor at an instrument tip sliding over porcine liver, across a simulated tumor (line 1) and over normal tissue (line 2). Reproduced with permission~\cite{hao2025high}. Copyright 2025, IEEE. iv) waveguide loss of a soft prosthetic hand pressing three tomatoes. Reproduced with permission~\cite{zhao2016optoelectronically}. Copyright 2016, The Authors, some rights reserved; exclusive licensee AAAS.}
\capline{c) Large-area contact and human–robot interaction: i) modular sensor strips that stop a collaborative robot on contact. Reproduced with permission~\cite{kim2026modular}. Copyright 2026, Elsevier. ii) tactile–proximity servoing of a robot end effector. Reproduced with permission~\cite{meng2025tactile}. Copyright 2025, IEEE. iii) Braille input with a fiber-optic tactile pad. Reproduced under the terms of the Optica Open Access Publishing Agreement~\cite{li2025integrating}. Copyright 2025, Optica Publishing Group. iv) five main types of physical interaction with the FBG-instrumented hand of a social robot; v) wavelength shifts of the four FBGs during a high-five and a handshake. Reproduced with permission~\cite{gaitanpadilla2025touch}. Copyright 2025, IEEE.}
\capline{d) Underwater and special environments: i) octopus-inspired soft arm searching for and grasping a bottle in a freshwater tank; ii) suction-cup signals and estimated contact direction during grasping. Reproduced under the terms of the CC BY 4.0 license~\cite{deldottore2026peripheral}. Copyright 2026, The Authors, published by Springer Nature. iii) ground demonstration of on-orbit service with an optical-fiber-knot sensor. Reproduced with permission~\cite{yu2023optum}. Copyright 2023, IEEE.}
\capline{e) Human-to-robot skill transfer with wearable FBG sensors: i) the sensors on a robotic hand for grasp-stability prediction and on a human operator's hand for demonstrations; ii) raw signals of the thumb sensor on the human and the robotic hand while grasping the same object. Reproduced with permission~\cite{xing2025taccap}. Copyright 2025, IEEE.}%
  }

\subsection{Large-Area Contact and Human–Robot Interaction}\label{sec:6.4}

Interaction through robot surfaces requires contact locations to be identified and motion states to be adjusted promptly. A Ruffini-type FBG skin was mounted on a robot surface to detect contact and temporarily stop motion during repeated execution of a taught trajectory. Modular optoelectronic strips were deployed on the links and wrist of a collaborative robot, where contact thresholds triggered the stopping of reciprocating motion (Figure~\ref{fig:6}c-i). Continuous encapsulation and reconfigurable modules provide different approaches to surface coverage. In these demonstrations, tactile signals were used primarily to trigger robot stopping. The control and actuation delays following contact detection, together with the distance required for the robot to stop, determine the practical effectiveness of such protective responses~\cite{massari2022functional,kim2026modular}. A dual-mode tactile–proximity optoelectronic sensor instead used pre-contact information for control. The same octagonal substrate carried 12 occlusion-based tactile elements and 5 infrared reflective proximity elements, enabling proximity servoing and tactile servoing separately or in combination to control the position and orientation of a robot arm's end effector (Figure~\ref{fig:6}c-ii)~\cite{meng2025tactile}.

Human–robot interaction tasks require different tactile output variables. Braille was recognized from spectral features by an optical-fiber MZI in a ring-shaped structure and from temporal signals by an optical-fiber ring resonator (Figure~\ref{fig:6}c-iii). A GaN multidirectional device used directional information for game input, and SLIMS reconstructed glove posture through color encoding. These systems converted non-imaging optical signals into text, direction or posture information, demonstrating the use of limited optical channels in different interaction tasks~\cite{shi2026multi,bai2020stretchable,li2025integrating,wang2025optical}. A flexible pressure-sensing device exploiting scattering in human tissue made the wearer's own tissue part of the optical path and estimated pressure using a convolutional model. Because the optical response was affected by wearing position and reattachment, recalibration using reference force data and model retraining were required after each reattachment to maintain the pressure estimation relationship~\cite{ando2026flexible}. A four-point FBG network on the hand of the social robot CASTOR was used to recognize nine classes of physical interaction, including high-fives, handshakes, slaps, strokes and pinches (Figure~\ref{fig:6}c-iv,v). Bagged decision trees achieved a test accuracy of 98\%, although recognition was weaker for localized, low-intensity gestures, such as pinching at FBG 2, which had a recall of 66.7\%~\cite{gaitanpadilla2025touch}. For human-to-robot skill transfer, a 2 × 3 plastic-optical-fiber fingertip array, 3 mm thick and weighing 1.98 g, used frustrated total internal reflection of cladding modes and could be worn on human fingertips or mounted on robotic ones. With a multilayer perceptron calibration it achieved sub-millimeter contact localization, a normal-force error of 0.424 N over 15 N and a 12.2 mN threshold, and gave consistent responses in synchronized human–robot teleoperation~\cite{chung2026poftac}.

\subsection{Underwater and Special-Environment Manipulation}\label{sec:6.5}

Underwater grasping requires holding an object after initial contact and then adjusting enveloping or pulling actions according to multipoint contact states (Figure~\ref{fig:6}d-i). An octopus-inspired soft arm embedded optoelectronic sensors in its suction cups, using three voltage outputs from each cup to detect contact and estimating contact direction through a weighted vector (Figure~\ref{fig:6}d-ii). When any local signal exceeded a threshold, suction by the corresponding pump group was triggered. The global control layer then combined the positions, number and directional information of suction cups in contact and, after the observation window ended, selected dorsal bending, ventral bending, twisting or continued adhesion. This approach first triggered adhesion through local contact feedback and then determined the tendon-driven strategy for the whole arm from multipoint contact information~\cite{deldottore2026peripheral}.

In a freshwater static tank, the soft arm demonstrated autonomous grasping of cylindrical objects at different inclination angles and contact locations, as well as smaller objects. It could also search for targets during slow movement and switch to grasping upon contact (Figure~\ref{fig:6}d-i). Local adhesion maintained contact while object orientation was assessed, allowing subsequent coordination of whole-arm motion. The system used embedded tactile information for contact assessment and action selection without external vision or remote-control input~\cite{deldottore2026peripheral}.

A soft-waveguide finger measured curvature and contact signals separately, using self-deformation as a reference to assess grasping states. The device underwent low-temperature and hydrostatic-pressure tests equivalent to a depth of 2700 m and demonstrated underwater pickup and release of a soft ball. These tests separately validated its sensing characteristics under specific environmental conditions and its signal readout capability during underwater operation~\cite{teeple2018soft}. An optical-fiber knot device intended for on-orbit service used repeated contact to identify screw protrusions on a solar panel after visual localization of a satellite, then removed the fasteners with a screwdriver to deploy the panel (Figure~\ref{fig:6}d-iii). This ground-based dual-arm demonstration combined visual localization, tactile search and tool operation, illustrating how contact information supplements local target identification and manipulation execution~\cite{yu2023optum}. An existing review specifically discusses functional materials, environmental adaptability and integration with robotic platforms for optical-fiber tactile sensing under extreme conditions, including electromagnetic interference, high temperatures, corrosive chemicals and ionizing radiation~\cite{tefery2026remote}.

\begin{table*}[!tbp]
\caption{Representative robotic applications and roles of tactile sensing}\label{tab:3}
\begin{tabular*}{\textwidth}{@{}>{\raggedright\arraybackslash}p{0.168\dimexpr\textwidth-8\tabcolsep\relax}>{\raggedright\arraybackslash}p{0.172\dimexpr\textwidth-8\tabcolsep\relax}>{\raggedright\arraybackslash}p{0.152\dimexpr\textwidth-8\tabcolsep\relax}>{\raggedright\arraybackslash}p{0.240\dimexpr\textwidth-8\tabcolsep\relax}>{\raggedright\arraybackslash}p{0.268\dimexpr\textwidth-8\tabcolsep\relax}@{}}
\toprule
\textbf{Representative system and task} & \textbf{Task information requirements} & \textbf{Measured signals} & \textbf{Action mechanism} & \textbf{Demonstrated results} \\
\midrule
Optical-waveguide soft prosthetic hand~\cite{zhao2016optoelectronically} & Surface shape and softness & Optical-loss responses during scanning and pressing & Predefined open-loop exploration, followed by measurement-based object selection & Surface and softness sensing; selection of a softer tomato \\
FBG force sensor at a laparoscopic instrument tip~\cite{hao2025high} & Stiffness differences in tissue under the instrument tip & Three-axis force from four FBG wavelength shifts & Sliding along predefined paths at a fixed indentation depth; force readout without feedback to the motion & Resultant force up to 1.66 N across a simulated tumor in porcine liver; 0.40–0.63 N over normal tissue \\
Optical-fiber knot grasping~\cite{pan2023knot} & Slip events during grasping & Friction-induced changes in optical intensity & Slip feedback triggers an increase in grip force & Feedback-on and feedback-off demonstrations with a progressively loaded cup \\
PapillArray grasping~\cite{wang2024robust} & Incipient slip and grasp stability & Tangential force time series from multiple pillars & Arm stopping and increase to a predefined grip force upon detection & Cross-device online deployment; 36 grasp-correction trials \\
FBG robotic hand~\cite{massari2019tactile} & Degree of contact and changes in object dimensions & Mean grating peak shift & Threshold-based stopping of finger closure; PID control maintains the peak-shift setpoint & One breakage in 20 nacho grasping trials; tracking of dimensional changes \\
Nine-waveguide soft gripper~\cite{fan2026intelligent} & Soft-finger bending, contact and slip & Multichannel optoelectronic voltages & Air-pump regulation through inner-channel voltage feedback; gripping parameters selected according to recognition results & Closed-loop grasping, slip adjustment and fruit-sorting demonstrations \\
FBG skin and modular sensing strips~\cite{massari2022functional,kim2026modular} & Surface contact and its location & Grating peak shifts or local reflected signals & Contact-triggered trajectory stopping & Online contact responses on robot surfaces \\
Underwater soft-waveguide finger~\cite{teeple2018soft} & Curvature and contact state & Intensity readings from separate waveguides & State readout during operation & Pressure-chamber and temperature characterization; underwater pick-and-place demonstration \\
Octopus-inspired underwater soft arm~\cite{deldottore2026peripheral} & Contact location, direction and object orientation & Three optoelectronic voltage outputs from each suction cup & Contact-triggered suction by pump groups; tendon-driven strategy selected using multipoint information & Autonomous grasping and motion-based search in a static water tank \\
Mechanoluminescent grasping~\cite{luo2026self} & Initial contact and slip events & Dynamic light peaks detected through photon counting & Light peaks trigger stopping, lifting or increased grip force & Mechanoluminescence used to trigger switching between robotic actions \\
Optical-fiber knot dual-arm manipulation~\cite{yu2023optum} & Locations of protruding fastening screws & Optical-fiber knot outputs from the pressure-sensing module & Tactile search after visual localization; tool change to remove fasteners & Ground-based solar-panel deployment demonstration \\
\bottomrule
\end{tabular*}
\begin{tablenotes}
Note: Action mechanisms are described according to the operations performed in the original studies. The prosthetic hand explored in open loop, and the underwater finger and the instrument-tip sensor were read out without feedback; in the other systems the tactile signal changed the robot action during the task. Property recognition, event detection and task outcomes represent capabilities at their respective stages; task outcomes retain the objects and operating conditions reported in the original studies.
\end{tablenotes}
\end{table*}

Table~\ref{tab:3} summarizes the task requirements, tactile observables and action responses of representative robotic applications. The main case studies illustrate how tactile information contributes to robotic decision-making, while other applications cover active exploration, contact-triggered stopping and interaction information readout.

Based on how tactile signals contribute to decision-making, the above examples can be divided into two categories according to their position in the perception–control chain. Most systems convert optical readout into threshold events or a single feedback variable to directly trigger predefined actions such as stopping, increasing grip force or activating suction. Examples include the mean peak-shift threshold of an FBG robotic hand, light-peak triggering in a mechanoluminescent gripper, contact-triggered stopping using modular sensing strips and local suction responses in an underwater soft arm. These reactive architectures are simple and offer controllable latency, but the available actions are manually predefined~\cite{luo2026self,deldottore2026peripheral,kim2026modular,massari2019tactile}. Another category introduces learning models to extract higher-level contact states from temporal or multichannel signals. Examples include PapillArray, which used a gated recurrent unit to recognize incipient slip and exclude interference from motion cessation, and a Ruffini-type FBG skin, which used a deep network to decode contact. Others are a flexible pressure-sensing device, which used a convolutional model for recalibration, and machine learning classification for texture and Braille recognition. In these systems, learning primarily serves perceptual decoding, and actions remain rule-triggered~\cite{massari2022functional,keser2021fiber,li2025integrating,wang2025optical,wang2024robust,ando2026flexible}. The wearable FBG finger sleeve TacCap used the same grating readout for human demonstrations and robot execution (Figure~\ref{fig:6}e-i). When the same object was grasped, the raw signals of the thumb sensors on the human and the robotic hand followed the same course, with smoother signals on the human hand (Figure~\ref{fig:6}e-ii). In a grasp-stability task using only tactile and proprioceptive inputs, training with demonstrations collected through the wearable device achieved a success rate of 90\%, compared with 10\% when FBG inputs were removed. In a skill-learning task incorporating vision, training solely on wearable demonstrations achieved a success rate of only 10\%~\cite{xing2025taccap}. Further integration of non-imaging optical tactile signals into policy learning or interaction prediction requires signal representations consistent across devices, interaction data acquired synchronously with robotic actions and online adaptation to calibration drift. Section~\ref{sec:7} discusses these issues in relation to calibration transfer and reliable large-area readout.

\section{Technical Challenges and Outlook}\label{sec:7}

Non-imaging optical tactile sensors have progressed from simple contact detection to multi-axis force estimation, distributed sensing and robotic feedback control. A central challenge for further development is obtaining accurate, physically interpretable tactile information consistently as contact conditions, mounting states and operating environments change. Addressing this challenge requires integrating mechanical load transmission, optical encoding, signal demodulation and robotic tasks while balancing improvements in device performance with measurement reliability, system scalability and practical deployment costs. Figure~\ref{fig:7} groups the challenges (a--c) and directions (d--f) into sensing and materials, readout and robotic tasks, which correspond to the three considerations set out in the Introduction. Sensing and materials determine whether contact variables are identifiable and whether identification holds over time and after remounting (Sections~\ref{sec:7.1} and~\ref{sec:7.2}). Readout sets spatial coverage, timing and integration at the system level (Sections~\ref{sec:7.3} and~\ref{sec:7.4}). Robotic tasks provide the evidence that links tactile measurements to task outcomes (Section~\ref{sec:7.5}).

\adrrfig{fig:7}{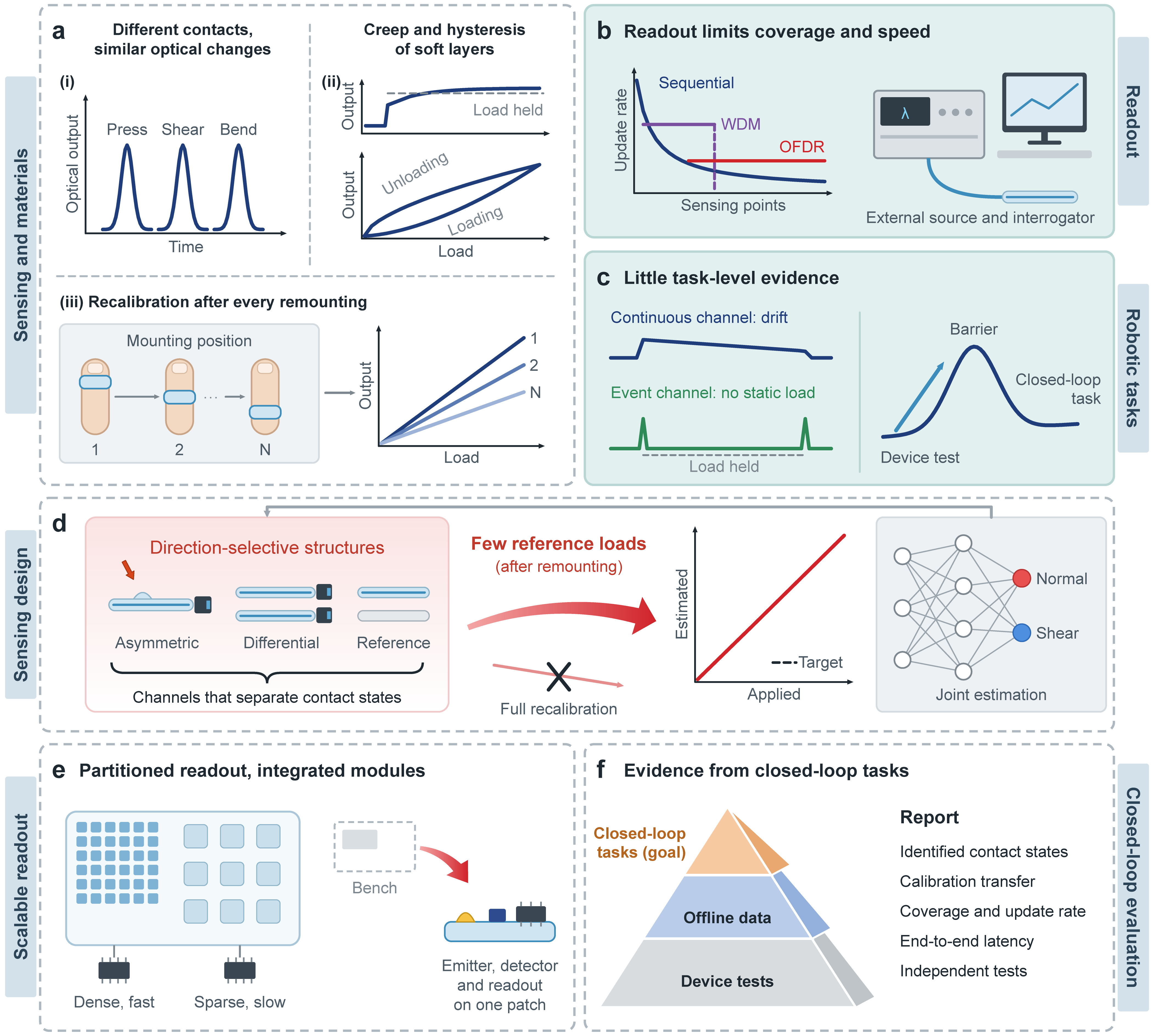}%
  {Technical challenges and directions for non-imaging optical tactile sensing.}%
  {%
\capline{Current challenges (a--c) and directions (d--f), grouped by sensing and materials, readout and robotic tasks (Sections~\ref{sec:7.1}--\ref{sec:7.5}). All plots are schematic and carry no values.}
\capline{a) Sensing and materials: i) pressing, shear and bending give similar changes in one optical channel; ii) soft layers creep under a held load and show hysteresis between loading and unloading; iii) the output--load relation changes with the mounting position, so the sensor is recalibrated after every remounting.}
\capline{b) Readout: the update rate of sequential readout falls as sensing points are added; wavelength-division multiplexing (WDM) is limited by its spectral budget; optical frequency-domain reflectometry (OFDR) reads many points at a lower rate; light sources and interrogators remain external.}
\capline{c) Robotic tasks: a continuous channel drifts and an event channel misses static loads; few studies go beyond device tests to closed-loop tasks.}
\capline{d) Sensing design: asymmetric, differential and reference channels that separate contact states; calibration updated from a few reference loads instead of a full recalibration; joint estimation of normal and shear forces checked against the applied loads.}
\capline{e) Scalable readout: dense, fast zones and sparse, slow zones read locally; emitters, detectors and readout integrated on one patch instead of bench instruments.}
\capline{f) Closed-loop evaluation: evidence moves from device tests and offline data to closed-loop tasks (Section~\ref{sec:6.1}); reports include the identified contact states, calibration transfer, coverage and update rate, end-to-end latency and independent tests.}%
  }

\subsection{Identifiability of Contact Information and Multiparameter Decoupling}\label{sec:7.1}

Optical tactile measurement is fundamentally an inverse problem in which contact states are inferred from optical observables. Normal force, tangential force, contact location and self-deformation may produce similar changes in optical intensity or spectra (Figure~\ref{fig:7}a-i), while temperature and mounting boundary conditions may further alter these responses. High optical sensitivity therefore does not guarantee that contact variables can be distinguished. Existing studies have obtained direction-dependent responses through asymmetric microfiber structures and distinguished temperature from pressure using optical features in different wavelength bands, indicating that coordinated design of mechanical structures and optical observation methods can improve multiparameter identifiability~\cite{xie2024optical,fan2026intelligent,yao2022single}.

Future research should further clarify the relationships between different contact variables and observed features and examine their stability under combined loading conditions. Direction-selective force-transmission structures (Figure~\ref{fig:7}d), differential observations and reference channels can be used to enhance target responses. Contact area, preload and body posture should be incorporated into calibration conditions. Existing research that corrected normal and tangential forces according to contact-object size demonstrates that contact geometry is an important factor affecting measurement mappings. Corresponding evaluations should extend beyond sensitivity under single-axis loading to include cross-axis errors under combined loads, joint estimation errors and noise sensitivity, thereby establishing the effective operating range of decoupling methods~\cite{xie2024optical,chen2023polymer}.

\subsection{Material and Interface Stability and Calibration Transfer}\label{sec:7.2}

Flexible materials and their encapsulation interfaces both transmit loads and contribute to optical-path modulation. Elastomer viscoelasticity, interfacial adhesion conditions and changes in optical connections may cause the same external load to produce different outputs over time; the output creeps under a held load and shows hysteresis between loading and unloading (Figure~\ref{fig:7}a-ii). Environmental effects also depend on the structure. For example, soft-waveguide curvature and contact-force sensing elements exhibited different response changes under low-temperature conditions, so environmental adaptability depends on the material and force-transmission mechanism. Existing underwater optoelectronic sensors have demonstrated stable sensing under dry and wet conditions, but their performance after prolonged immersion and repeated use still requires dedicated evaluation~\cite{teeple2018soft,deldottore2026peripheral}. Emission-based materials also age under repeated loading: a CaZnOS:Nd,Er mechanoluminescent fiber retained about 67\% of its initial intensity after 4200 cycles at 2 N~\cite{zheng2025flexible}. Medical use adds sterilization, which one enclosed FBG instrument withstood at 180 °C~\cite{li2025high}.

Improving stability requires simultaneous advances in materials, interfaces and correction methods. Material and encapsulation design should address loading history, interfacial slip and connection reliability, while signal processing can use reference channels, baseline normalization and temperature compensation to reduce identifiable disturbances. On this basis, model updates using small amounts of standard contact data could be explored to reduce the calibration burden after remounting (Figure~\ref{fig:7}d). Accurate estimation after recalibration and direct transfer of an existing model are distinct capabilities (Figure~\ref{fig:7}a-iii). For example, a reflective optical sensor mounted on the back of a finger compensated for mounting and interindividual differences through reference data collection and CNN retraining after each attachment. Future studies should report calibration data requirements, update times and independent test errors after transfer to clarify the conditions under which simplified correction or complete recalibration is appropriate~\cite{wang2024robust,ando2026flexible}.

\subsection{Spatial and Temporal Constraints of Large-Area Sensing}\label{sec:7.3}

Scaling from local sensing elements to large-area tactile skins simultaneously increases the demands on spatial encoding, signal transmission and real-time processing. Different readout methods impose different constraints. FBG wavelength-division multiplexing requires sufficient spectral space for individual sensing elements and their wavelength shifts. Sequential optoelectronic sampling can reduce connection complexity through shared wiring, but increasing the number of nodes extends the full-array readout cycle (Figure~\ref{fig:7}b). OFDR provides spatial localization information along optical fibers. Its update rate is limited by along-fiber acquisition, correlation-based demodulation and subsequent inference; one complete acquisition of a microfiber skin took about 1 s~\cite{li2026continuous}. Coverage area, spatial resolution and update rate should therefore be evaluated together within the complete readout architecture~\cite{heo2006tactile,li2026continuous,kim2026modular}.

Spatial scaling also requires examining whether information from simultaneous multipoint contacts can be separated and how mechanical coupling between neighboring regions affects localization and force estimation. Future research could explore non-uniform sensing layouts, partitioned acquisition, local preprocessing and on-demand demodulation according to the sensing requirements of different robot regions, expanding coverage while maintaining response speed in critical areas (Figure~\ref{fig:7}e). Performance reports should clearly distinguish sensing-element spacing, spatial resolution of optical demodulation, localization errors of learning models and the ability to distinguish neighboring contacts, together with the loading and test conditions.

\subsection{Optoelectronic System Integration and Manufacturing Consistency}\label{sec:7.4}

Miniaturization of sensing elements needs to proceed alongside the integration of light sources, detectors, coupling interfaces and demodulation circuits. For approaches using external optical instruments, the size, cost and connection requirements of the complete system may become major constraints on robotic deployment (Figure~\ref{fig:7}b). For approaches with locally integrated optoelectronic devices, mechanical compatibility between rigid chips and flexible contact layers must be addressed. Monolithically integrated GaN emitting and detecting devices, together with modular LED–photodiode platforms (Figure~\ref{fig:7}e), have demonstrated feasible approaches to reducing the size of local readout units and simplifying system connections~\cite{li2026continuous,luo2025optical,kim2026modular}. Optical fibers can also deliver actuation energy to an end effector. A photothermal hydrogel microgripper fabricated by two-photon polymerization at an optical-fiber end face measured only 38×38×61 μm³ and was opened and closed by an 808 nm laser delivered through the fiber, with a peak gripping force of approximately 2.2 μN per jaw. The device itself had no force or tactile readout, and its gripping force had to be measured using an externally calibrated microspring probe. If actuation energy and tactile signals could be transmitted simultaneously through the same optical fiber, the number of end-effector connections could be reduced further~\cite{pan2026optical}. A tactile gripper on a single-mode fiber has combined sensing and actuation at the microscale: a two-photon-polymerized SU-8 wave spring acted as a Fabry–Pérot force sensor beside a pH-responsive gripper. The system sorted micro-objects and measured the Young's modulus of zebrafish zygotes, although its actuation was chemical rather than optical~\cite{li2025tactile}.

Further integration should account for optical alignment, packaging tolerances, connection durability and consistency in mass manufacturing. Variations in sensitive-layer thickness, waveguide geometry and device positioning may increase cross-device calibration costs, so manufacturing process control and calibration methods need to be designed together. System energy consumption should cover the complete information acquisition chain. For ZnS:Cu/PDMS mechanoluminescent devices such as SOFTS, mechanical excitation eliminates the need for external optical excitation, but photodetection, signal acquisition, computation and communication must still be included in the energy budget. Future studies should clearly define system boundaries and report the dimensions, power consumption, update rate and maintenance requirements of complete modules, allowing the benefits of miniaturization and low power consumption to be assessed under practical operating conditions~\cite{luo2026self}.

\subsection{Coordinated Sensing and Evaluation for Robotic Tasks}\label{sec:7.5}

Robotic tasks differ substantially in their requirements for tactile information. Sustained grasping requires stable tracking of contact states during the holding phase, incipient slip detection emphasizes transient information and its timeliness, and object property estimation additionally depends on motion and contact conditions during active exploration. Mechanoluminescence-triggered control, FBG grasping feedback and optical-fiber knot slip detection illustrate different ways in which optical observations contribute to manipulation. For devices that primarily output dynamic luminescence, the ability to characterize static loads should be validated separately. For continuous readout approaches, low-frequency drift and responses to rapid events need to be evaluated together (Figure~\ref{fig:7}c)~\cite{luo2026self,pan2023knot,massari2019tactile}. Most studies reviewed here report device characterization or offline recognition (Table~\ref{tab:2}), and fewer let the tactile signal change a robot action during a task (Table~\ref{tab:3}).

Combining continuous load channels with rapid event channels is a promising route to coordinated tactile feedback. Synchronized measurements and contact-state estimation could support grasp maintenance, slip correction and active exploration across different timescales. Existing studies of learning-based slip detection and multimodal waveguide grippers have shown that tactile information can contribute to robotic feedback regulation. Future evaluations should cover end-to-end latency arising from acquisition, filtering, inference, communication and actuator response, and include false positives, missed detections, grasp success rates and object damage in the analysis (Figure~\ref{fig:7}f)~\cite{fan2026intelligent,wang2024robust}. One sensor combined force and tactile sensing in a single device, fusing FBG-based three-axis force measurement (RMSE of 0.010–0.022 N) with a camera-based contact image used for depth reconstruction~\cite{liang2026fvis}. Processing can begin in the sensing material itself: in a mechano-optical synapse, mechanoluminescence was stored in an adjacent photostimulated-luminescence layer and later read out with 980 nm light, supporting in-sensor computing for visual–tactile perception~\cite{guo2025sensor}.

To improve reproducibility, studies should also clearly distinguish the data and conditions used for device calibration, perception model training and robotic task validation. Independent tests should be organized by device, object and mounting batch, with controlled comparisons conducted under identical hardware and control conditions. A review of machine learning for tactile perception noted that standardized measurements exist for sensor quality, but no benchmark has yet been established for comparing the overall performance of sensors and algorithms~\cite{hu2023machine}. Linking contact mechanics, optical observations and task performance is necessary to determine the operating conditions under which sensors can provide reliable information and how that information translates into verifiable improvements in manipulation.

\subsection{Conclusions}\label{sec:7.6}

Non-imaging optical tactile sensing converts contact-induced changes in light propagation or emission into information for robotic perception and control. Three considerations recur across the mechanisms reviewed. The first is whether the target contact variables remain identifiable under the relevant contact and calibration conditions, and the second is whether spatial coverage and feedback timing hold at the system level. The third is what level of evidence, from device characterization to closed-loop tasks, supports a reported capability. The studies reviewed here support a task-driven design principle: mechanical structures, optical encoding and readout should be selected together to identify the contact states required by the robot and deliver them within the feedback timescale. The mechanical structure determines whether these quantities produce distinguishable optical changes, through load paths, asymmetric or differential layouts and reference channels. Optical encoding and readout set the number of independent channels, the update rate and the hardware that must sit at the contact site or can be placed remotely. Table~\ref{tab:routes} compares these properties across the technical routes. Identification must then hold under combined loads, temperature changes and remounting, and the end-to-end feedback time, from a change in contact state to the actuator response, must match the task, as the applications in Table~\ref{tab:3} illustrate. Applying this principle more widely requires reliable multiparameter estimation, stable materials and interfaces, efficient calibration transfer and scalable optoelectronic integration. Coupled with independent testing, comparable metrics and closed-loop task evaluation, this review provides a principled design and evaluation framework for future research in robotic tactile manipulation.

\subsection*{Acknowledgements}

This work was supported in part by the National University of Singapore under the NUS Start-up Grant (FY2026). The authors thank Bingsheng Xie, Enmai Gaozhang and Yujie Zeng for their assistance.

\subsection*{Author Contributions}

Hao Wu and Yu Feng contributed equally to this work. They collected and curated the literature, drafted the manuscript, prepared the figures and tables, and revised and proofread the manuscript. Haotian Guo cross-checked the literature and manuscript and contributed to proofreading and revision. Huichan Zhao, Changsheng Wu and Jianshu Zhou supervised the work, guided the scope and structure of the review, and critically reviewed and revised the manuscript.

\subsection*{Conflict of Interest}

The authors declare no conflict of interest.

\subsection*{Data Availability Statement}

Data sharing is not applicable to this article as no new data were created or analyzed in this study.

\bibliography{references}

\end{document}